\documentclass[preprint,12pt]{elsarticle}
\usepackage[symbol]{footmisc}
\def\correspondingauthor{\footnote{Corresponding author.
E-mail address:abdelouahed\_laazoufi@um5.ac.ma}}

\usepackage{amssymb}
\usepackage{stmaryrd}
\usepackage{amsmath}
\usepackage{multirow} 
\usepackage{xcolor}   
\usepackage{amsmath}  

\usepackage{amsmath}
\usepackage{subcaption}
\usepackage{float}
\usepackage{multirow}
\usepackage{verbatim}
\usepackage{xcolor}
\usepackage{tikz}
\usepackage{amsmath}
\usepackage{multirow} 
\usepackage{xcolor}   
\usepackage{graphicx} 
\usepackage{subcaption} 
\usetikzlibrary{arrows.meta, positioning, shapes.geometric, fit, calc}
\journal{}

\begin{document}

\begin{frontmatter}



\title{Point Cloud Quality Assessment Using the  Perceptual Clustering Weighted Graph (PCW-Graph) and  Attention Fusion Network}

\author{Abdelouahed Laazoufi$^{a,}$\correspondingauthor{},Mohammed El Hassouni$^{a}$ and Hocine Cherifi$^{b}$} 
\affiliation{organization={FLSH, FSR, LRIT},
            addressline={
Mohammed V University in Rabat}, 
            country={Morocco}}
            
\affiliation{organization={ICB UMR 6303 CNRS},
            addressline={Université Bourgogne Europe}, 
            city={Dijon},
            postcode={21000}, 
            country={France}}
\begin{abstract}
No-Reference Point Cloud Quality Assessment (NR-PCQA) is critical for evaluating 3D content in real-world applications where reference models are unavailable. Existing NR-PCQA methods often fail to capture structural and perceptual relationships, leading to inconsistent quality predictions. To address this issue, we propose a Perceptual Clustering Weighted Graph (PCW-Graph) that models point cloud distortions through a structured learning process.

Our approach follows three main steps: (1) Perceptual Feature Extraction - Color, curvature, and saliency features are computed to capture geometric and visual distortions. (2) Graph-Based Representation — Perceptual clusters form nodes, and weighted edges encode feature similarities, ensuring spatial and structural consistency. (3) Adaptive Quality Prediction - A Graph Attention Fusion Network (GAF) dynamically refines feature importance and estimates quality scores via regression to improve correlation with human perception.

We evaluate the proposed method on three benchmark datasets: WPC, SJTU PCQA, and ICIP2020. It achieves high Pearson (PLCC $\geq 0.93$) and Spearman (SRCC $\geq  0.91$) correlation coefficients. These results show better alignment with human perceptual scores than alternative NR-PCQA methods. The approach also reduces Root Mean Square Error (RMSE) by up to 15\%. This confirms its effectiveness and robustness for blind point cloud quality assessment.

\end{abstract}




\end{frontmatter}



\section{Introduction}
\label{sec1}
The adoption of 3D models has recently grown significantly across multiple fields, such as architecture, cultural heritage preservation, virtual and mixed reality, and computer-aided diagnostics. However, processes like compression and simplification applied to these 3D models can introduce distortions that degrade the visual quality of 3D point clouds. This issue has increased the need for effective methods to evaluate perceived quality. In the past, distortion levels in 3D models were assessed by human observers, a process that is time-consuming and resource-intensive. To improve efficiency, objective methods have been introduced as a solution \cite{mohammadi2014subjective}. These methods employ automatic measurements that simulate the decisions made by an optimal human observer. These metrics can be categorized into three types: Full Reference (FR) \cite{wang2004image,tian2017geometric,meynet2020pcqm,yang2020inferring,chen2021layered}, Reduced Reference (RR) \cite{viola2020reduced,abouelaziz2015reduced}, and No Reference (NR) \cite{liu2021pqa,mittal2012no,zhang2015feature,liu2023point,zhang2022no,abouelaziz2016no,abouelaziz2020no}. Among these, blind methods, which do not rely on reference models, are gaining increasing importance in real-world applications \cite{lin2019blind,abouelaziz2016curvature,abouelaziz2017convolutional,abouelaziz20203d}.

A 3D point cloud (PC) consists of a collection of points defined by geometric coordinates and potentially other attributes such as color, reflectance, or surface normals. Three-dimensional point clouds contain information about the shape and geometry of an object, with points irregularly arranged in space compared to images and videos. Therefore, effective methods for feature extraction from these scattered points must be explored to assess quality accurately.

Currently, very few metrics are available for No-Reference Point Cloud Quality Assessment (NR-PCQA). ResSCNN \cite{liu2023point} proposes using voxelization to merge nearby points into a single voxel. Chetouani et al. \cite{chetouani2021deep} take a different approach by extracting hand-crafted features at the patch level and using a traditional CNN model for quality regression. PQA-net \cite{tian2017geometric} and IT-PCQA \cite{yang2022no} employ multiview projection for feature extraction. Zhang et al. \cite{zhang2022no} adopt a unique strategy, utilizing statistical distributions to estimate quality-related parameters from the geometry and attribute color distributions. Fan et al. attempt to estimate the visual quality of point clouds by analyzing captured video sequences \cite{fan2022no}. Liu et al. \cite{liu2023point} utilize an end-to-end sparse CNN to predict quality. Yang et al. \cite{yang2022no} further advance this field by transferring quality scores from natural images to point cloud rendering images using domain adaptation techniques.

However, in CNNs, the gradient of one kernel point's weight during backpropagation is only related to the spatially corresponding isolated point in the patch due to the element-wise multiplication between the convolution kernel and the patch. This limited point interaction hampers the thorough extraction of structural and textural features \cite{liu2019relation}.

Current no-reference metrics often ignore the rotation invariance, shift, and scale of point clouds, leading to inconsistent objective ratings for the same sample under different scales or positions. This oversight contrasts with most PCQA subjective studies, where observers have the freedom to choose the viewing angle and point cloud position \cite{liu2023point,yang2020predicting,liu2021reduced,perry2020quality}. Ignoring these invariances can reduce the robustness of the metric.

Recently, graphs have received significant attention for point cloud processing applications \cite{yang2013saliency,sakiyama2019eigendecomposition,zeng20193d}. Much of this progress is attributed to the ability of graph-based representations to model high-dimensional visual data, such as 3D point clouds, by implicitly capturing local neighbor relationships and highlighting their importance. Graphs are highly flexible data structures, making them ideal for representing 3D point clouds in an interpretable manner. The connectivity between a vertex and its neighboring vertices provides valuable insights into the perceptual characteristics of a point cloud. The application of graph representations has demonstrated significant efficacy in addressing various computer vision tasks, including image segmentation \cite{mourchid2019general} and classification \cite{ribas2022complex}. El Hassouni et al. \cite{el2022learning} use graph feature learning and random forest regression to evaluate the visual quality of colored meshes.

The key contributions of this work are outlined as follows:
\begin{itemize}
\item The paper proposes a new no-reference point cloud quality assessment (PCQA) method called the Perceptual Clustering Weighted Graph (PCW-Graph).
\item The method begins by segmenting distorted point clouds into regions through clustering based on perceptual features. The clusters serve as nodes in the graph, and the ties between nodes are represented by weights that reflect the similarity of the cluster centers.
\item A Graph Attention Fusion Network (GAF) is employed during model training to better represent features, understand connections between perceptual features, and allow for flexible weight adjustments.
\item We compare the proposed method with several full-reference and state-of-the-art No-Reference PCQA metrics.
\end{itemize}

The structure of this paper is as follows: Section 2 describes the proposed method, Section 3 details the experimental results, and Section 4 concludes the paper.

\section{Related work}
Based on the kind of data they use, existing PCQA methods are generally categorized into two types: projection-based and point cloud-based methods. 
\subsection{Point Cloud-Based Methods}
Point cloud-based methods utilize raw point cloud data directly, extracting geometric, color, and other features to evaluate quality. Mekuria and Tian et al. were among the first to introduce point-based quality assessment methods, namely PSNR\_p2po \cite{mekuria2016evaluation} and PSNR\_p2pl \cite{tian2017geometric}, marking the early development of PCQA methodologies. Later, PSNR\_yuv \cite{mekuria2017performance} was introduced to assess texture distortion in color point clouds. Alexiou et al. \cite{alexiou2018point} used the angle between corresponding points to characterize the degradation of distorted point clouds. Meynet et al. \cite{meynet2019pc} proposed the MSDM method, which relies on local curvature statistics to measure point cloud quality. Javaheri et al. \cite{javaheri2020generalized} employed the generalized Hausdorff distance in their PCQA work.\\
In addition to geometric features, color information has received significant attention. Meynet et al. \cite{meynet2020pcqm} enhanced the performance of MSDM by incorporating color properties. Viola et al. \cite{viola2020color} used global color statistics to measure point cloud distortion levels. In contrast, Alexiou et al. \cite{alexiou2020towards} calculated a structural similarity index in feature space based on both geometric and color characteristics. Diniz et al. \cite{diniz2020towards,diniz2020multi,diniz2020local} proposed a technique that utilizes local luminance pattern (LLP) and local binary pattern (LBP) descriptors to extract statistical data from point clouds. Yang et al. \cite{yang2020inferring,yang2022mped} leveraged the multi-scale potential energy differential (MPED) of point clouds in conjunction with local graph representations to assess quality scores.\\
The Reduced Reference (RR) approach is particularly useful when a complete reference point cloud is unavailable. Viola et al. \cite{viola2020reduced} introduced a new RR measure that uses a linear optimization algorithm to determine the optimal combination of features, including geometry, brightness, and normals. Liu et al. \cite{liu2021reduced} proposed an RR method for predicting the quality of V-PCC compressed point clouds based on geometric and color information. Similarly, Su et al. \cite{su2023support} developed an RR point cloud quality assessment method based on Support Vector Regression (SVR).\\
Despite these advancements, it is important to note that reference point clouds are not always accessible in real-world applications. Zhang et al. \cite{zhang2022no} introduced a No-Reference (NR) method that utilizes 3D natural scene statistics (3D-NSS) to extract color and geometric features, with SVR used to predict quality. Zhou et al. \cite{zhou2024blind} proposed a structure-guided resampling-based blind quality assessment method for point clouds.\\
The integration of deep learning has significantly advanced PCQA. Chetouani et al. \cite{chetouani2021deep} used a deep neural network (DNN) to map feature representations to quality scores. Liu et al. \cite{liu2023point} presented an NR-PCQA technique using sparse convolutional neural networks (SCNNs). Shan et al. \cite{shan2023gpa} introduced a Multi-task Graph Convolutional Network (GPA-Net) for NR-PCQA. Wang et al. \cite{wang2023non} proposed a non-local geometric and color gradient aggregation graph model with a multi-task learning module. Tliba et al. \cite{tliba2023efficient} suggested a PCQA measure based on a Dynamic Graph Convolutional Neural Network (DGCNN). Su et al. \cite{su2023bitstream} developed a bitstream-based neural network model to assess the perceptual quality of point clouds. Liu et al. \cite{liu2022no} introduced a new NR-PCQA bitstream model for evaluating the quality of V-PCC encoded point clouds.
\subsection{Projection-Based Methods}
Projection-based techniques evaluate quality by projecting the point cloud onto a 2D image plane and analyzing the resulting images. This allows the application of existing Image Quality Assessment (IQA) techniques, such as SSIM \cite{wang2004image}, MS-SSIM \cite{wang2003multiscale}, IW-SSIM \cite{wang2010information}, and VIFP \cite{sheikh2006image}. Hua et al. \cite{hua2021bqe} proposed a blind assessment method that extracts geometric, color, and joint properties by projecting the point cloud onto a plane. Freitas et al. \cite{freitas2023point} introduced a Full-Reference (FR) PCQA measure that combines texture and geometry characteristics in projected texture maps to estimate point cloud quality accurately. Zhou et al. \cite{zhou2023reduced} developed an RR-PCQA method using content-based saliency projection.\\
No-reference (NR) approaches and deep learning techniques have also gained traction in projection-based methodologies. Tao et al. \cite{tao2021point} proposed a multi-scale feature fusion network to weight the quality scores of local patches within a graph. Liu et al. \cite{liu2021pqa} employed a DNN to predict quality scores and probability vectors by extracting multi-view features, inspired by multi-task models. Tu et al. \cite{tu2022v} introduced a two-stream CNN designed to extract features from both texture and geometric projection maps. Yang et al. \cite{yang2022no} incorporated unsupervised adversarial domain adaptation into PCQA to transfer knowledge from the source domain to the target domain. Recently, Xie et al. \cite{xie2023pmbqa} proposed a projection-based NR-PCQA metric using multi-modal learning and graph structures for multi-modal feature fusion. Zhang et al. \cite{zhang2024gms} developed a multi-projection grid mini-patch-based method (GMS-3DQA) to evaluate 3D model quality while reducing resource consumption. Zhang et al. \cite{zhang2022mm} extended single-modal approaches by retrieving multi-modal data (MM-PCQA) from 3D point clouds to improve perceptual quality.
\subsection{Challenges in PCQA Methods}
\begin{itemize}
\item Point Cloud-Based Approaches: These methods enable a more comprehensive understanding of geometric and spatial characteristics. However, their effectiveness largely depends on the quality of feature extraction techniques. Inadequate or inaccurate feature extraction can lead to biased evaluations.
\item Projection-Based Approaches: A significant limitation of these methods is the inevitable loss of data during the projection process. This loss can impact the accuracy of quality assessment, especially if critical details are omitted. Additionally, the choice of views and angles in the projection can influence the quality of the resulting images, potentially leading to inconsistencies in evaluation outcomes.
\end{itemize}

\section{The proposed method}
\begin{figure*}[htbp]
\centering
\includegraphics[width=1.0\textwidth]{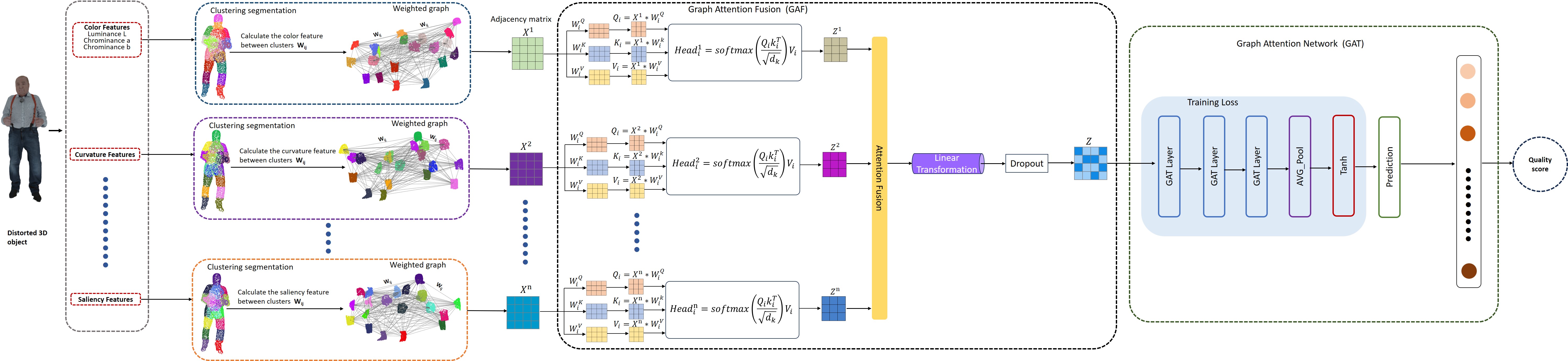}
\caption{The general framework of the proposed method.}
\label{fig:our_model}
\end{figure*}

The framework of our proposed method is illustrated in Figure \ref{fig:our_model}, comprising four key modules: the feature extraction module, graph construction module, graph attention fusion network (GAF) module, and quality regression module. First, perceptual features are extracted from each distorted point cloud. These features are then utilized to construct a weighted graph by partitioning the distorted point cloud into distinct regions, where each node corresponds to a cluster, and edges between nodes are weighted to reflect the similarity between their respective cluster centers. The adjacency matrix of this weighted graph enables the integration of multiple features for assessing the visual quality of the point cloud. Subsequently, the graph attention fusion network (GAF) processes the graph to generate view-specific feature embeddings and fuses them with graph embeddings. Finally, multi-graph fusion is applied to aggregate information from multiple GAF-generated graphs, each representing different feature types. The fused output is fed into a Graph Attention Network (GAT) to predict the final perceptual quality score of the point cloud.

\subsection{Perceptual Features Extraction}\label{subsec1}  
For each point cloud \(\mathbf{PC} = \{ \mathbf{p}_i \}_{i=1}^N \), we extract a set of perceptual features \(\mathbf{F}_{\text{perc}}\) defined as:  

\begin{equation}  
\mathbf{F}_{\text{perc}} = \mathcal{F}_{\text{perc}}(\mathbf{PC})  
\end{equation}  

where \(\mathcal{F}_{\text{perc}}\) denotes the perceptual feature extraction function. The selected features include color, curvature, and saliency.  

\begin{itemize}  
\item \textbf{Color:} Color plays a critical role in perceptual quality assessment. While most 3D models represent color using RGB channels, this space is not well-aligned with human perception. To address this, we project RGB values into the CIELAB color space using the method in \cite{nehme2020visual}, which is widely adopted for quality evaluation tasks.  

\item \textbf{Curvature:} Curvature quantifies the deviation of a surface from flatness at a given point. For each point \(\mathbf{p}_i\), we compute a covariance matrix \(\mathbf{C}\) over its local neighborhood \(\mathcal{N}_i\) with radius \(r\):  
\begin{equation}  
\mathbf{C} = \frac{1}{|\mathcal{N}_i|} \sum_{\mathbf{p}_j \in \mathcal{N}_i} (\mathbf{p}_j - \bar{\mathbf{p}})(\mathbf{p}_j - \bar{\mathbf{p}})^\top  
\label{eq:covariance}  
\end{equation}  
where \(|\mathcal{N}_i|\) is the neighborhood size, \(\mathbf{p}_j\) are neighboring points, and \(\bar{\mathbf{p}}\) is their centroid. Let \(\lambda_1 \geq \lambda_2 \geq \lambda_3\) be the eigenvalues of \(\mathbf{C}\). The curvature is computed as:  
\begin{equation}  
\text{Curv}(\mathbf{p}_i) = \frac{\lambda_3}{\lambda_1 + \lambda_2 + \lambda_3}  
\label{eq:curvature}  
\end{equation}  

\item \textbf{Saliency:} Visual saliency identifies regions that attract human attention. To compute saliency for a point \(\mathbf{p}_i\), we first generate two Gaussian-smoothed versions \(\mathbf{g}_1(\mathbf{p}_i)\) and \(\mathbf{g}_2(\mathbf{p}_i)\) at scales \(\sigma_1\) and \(\sigma_2\) (\(\sigma_1 < \sigma_2\)). The saliency is then:  
\begin{equation}  
\text{Sal}(\mathbf{p}_i) = \left| \mathbf{n}_i \cdot \left( \mathbf{g}_1(\mathbf{p}_i) - \mathbf{g}_2(\mathbf{p}_i) \right) \right|  
\label{eq:saliency}  
\end{equation}  
where \(\mathbf{n}_i\) is the surface normal at \(\mathbf{p}_i\), and \(| \cdot |\) denotes the absolute value.  
\end{itemize}  

\subsection{PC segmentation by clustering}\label{subsec2}

Clustering groups objects into clusters based on similarity without prior knowledge of data structure. In point cloud quality assessment, this step identifies regions with shared geometric or perceptual properties within \(\mathbf{PC} = \{ \mathbf{p}_i \}_{i=1}^N \), enabling localized distortion analysis and spatial dependency modeling. Robust clustering improves the alignment of quality metrics with human perception by capturing contextual relationships between points. Common techniques include Spectral Clustering \cite{ma2010point}, Mean Shift \cite{comaniciu2002mean}, Watershed \cite{yang2020individual}, DBSCAN \cite{deng2020dbscan}, KD-Tree \cite{guo2023kd}, and K-means \cite{davidson2002understanding}. Among these, K-means \cite{faber1994clustering,maulik2000genetic} remains the most widely adopted partitioning algorithm.

The K-means workflow involves:
\begin{enumerate}
    \item Selecting the number of clusters \(k\),
    \item Initializing cluster centroids \(\mathbf{t}_i\),
    \item Assigning points \(\mathbf{p}_s\) to the nearest centroid via the Nearest-Neighbor rule \cite{wang2006genetic},
    \item Updating centroids iteratively until convergence.
\end{enumerate}
Convergence is determined by minimizing the error function:
\begin{equation}
    \mathcal{E_r} = \sum_{i=1}^{k} \sum_{\mathbf{p}_s \in \mathcal{C}_i} \left\| \mathbf{p}_s - \mathbf{t}_i \right\|^2,
    \label{eq:error}
\end{equation}
where \(\mathcal{C}_i\) denotes the \(i\)-th cluster and \(\mathbf{t}_i\) its centroid. Smaller \(\mathcal{E_r}\) values indicate higher intra-cluster similarity.

We partition the deformed point cloud into clusters using K-means on perceptual features \(\mathbf{F}_{\text{perc}}\). These clusters then define nodes in a graph representation for subsequent quality analysis.
\subsection{Perceptual
Clustering Weighted Graph (PCW-Graph) construction}\label{subsec3}
The structural relationships between perceptual clusters are encoded in a weighted graph \(\mathcal{G} = (\mathcal{V}, \mathcal{E}, \mathbf{W})\), where \(\mathcal{V}\) represents cluster nodes, \(\mathcal{E}\) denotes their connections, and \(\mathbf{W} \in \mathbb{R}^{k \times k}\) contains edge weights based on both geometric proximity and perceptual similarity. This graph construction process comprises four sequential stages that transform clustered features \(\mathbf{F}_{\text{perc}}\) into relational representations suitable for quality assessment.

The graph nodes \(\mathcal{V} = \{\mathcal{C}_1, \dots, \mathcal{C}_k\}\) correspond to the \(k\) clusters identified through K-means segmentation (Section~\ref{subsec2}), where each cluster \(\mathcal{C}_i\) groups points with similar perceptual characteristics. Spatial adjacency between clusters determines the edge set \(\mathcal{E}\), with two clusters \(\mathcal{C}_i\) and \(\mathcal{C}_j\) being connected \((i,j) \in \mathcal{E}\) if they lie within a neighborhood radius \(r\) in the 3D space. The weight matrix \(\mathbf{W}\) quantifies interaction strengths using a dual metric that combines geometric distance and perceptual feature similarity.

\begin{equation}
    \mathbf{W}_{ij} = \begin{cases}
        \text{Sim}(\mathcal{C}_i, \mathcal{C}_j) \cdot D_e(\mathcal{C}_i, \mathcal{C}_j) & \text{if } \mathcal{C}_j \in \mathcal{N}_r(\mathcal{C}_i) \\
        0 & \text{otherwise}
    \end{cases}
    \label{eq:weight_matrix}
\end{equation}

The construction begins with cluster centroid computation. For each cluster \(\mathcal{C}_i\), we calculate its perceptual centroid \(\boldsymbol{\mu}_i \in \mathbb{R}^d\) by averaging the feature vectors of its constituent points:

\begin{equation}
    \boldsymbol{\mu}_i = \frac{1}{|\mathcal{C}_i|} \sum_{\mathbf{p}_k \in \mathcal{C}_i} \mathbf{F}_{\text{perc}}(\mathbf{p}_k),
    \label{eq:centroid}
\end{equation}

where \(|\mathcal{C}_i|\) denotes the number of points in cluster \(\mathcal{C}_i\), and \(\mathbf{F}_{\text{perc}}(\mathbf{p}_k)\) represents the perceptual features (color, curvature, saliency) of point \(\mathbf{p}_k\).

Geometric separation between clusters is quantified through Euclidean distance between their centroids:

\begin{equation}
    D_e(\mathcal{C}_i, \mathcal{C}_j) = \|\boldsymbol{\mu}_i - \boldsymbol{\mu}_j\|_2 = \sqrt{\sum_{d=1}^3 (\mu_i^{(d)} - \mu_j^{(d)})^2},
    \label{eq:euclidean_distance}
\end{equation}

where \(\mu_i^{(d)}\) indicates the \(d\)-th spatial coordinate (X, Y, or Z) of centroid \(\boldsymbol{\mu}_i\). This distance metric preserves the original point cloud geometry while operating at cluster resolution.

Perceptual similarity is then computed using a radial basis function (RBF) kernel that converts geometric distances into similarity scores:

\begin{equation}
    \text{Sim}(\mathcal{C}_i, \mathcal{C}_j) = \exp\left(-
    \frac{D_e(\mathcal{C}_i, \mathcal{C}_j)}{2\alpha^2}\right),
    \label{eq:similarity}
\end{equation}

The bandwidth parameter \(\alpha > 0\) controls the similarity decay rate - smaller values produce sharper similarity gradients, emphasizing local neighborhood relationships, while larger values permit longer-range interactions. Through experimental validation, we set \(\alpha = 0.15r\) where \(r\) is the neighborhood radius from Section~\ref{subsec2}.

The final weight matrix \(\mathbf{W}\) combines these metrics through element-wise multiplication of similarity scores and inverse distance scaling for connected node pairs. This dual-metric approach ensures strong weights between clusters that are both perceptually similar and geometrically proximate, while suppressing connections between distant or dissimilar clusters. The resulting graph structure preserves critical spatial-perceptual relationships while reducing computational complexity through cluster-level abstraction.
\begin{figure}[htbp]
    \centering
    \includegraphics[width=0.8\textwidth]{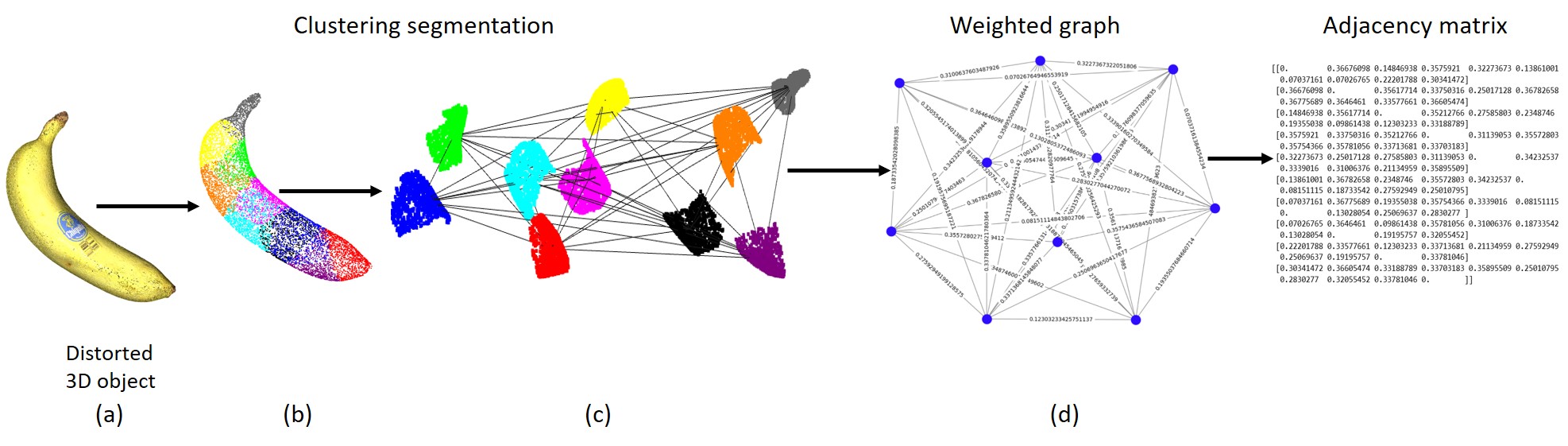}
    \caption{Weighted graph construction workflow: (a) Original point cloud, (b) Cluster segmentation using \(\mathbf{F}_{\text{perc}}\), (c) Centroid computation (Eq.~\ref{eq:centroid}), (d) Edge weighting via Eq.\ref{eq:weight_matrix}.}
    \label{fig:clust_seg}
\end{figure}
\subsection{Attention Mechanism}\label{subsec4}
The attention mechanism processes the clustered graph structure through its adjacency matrix \(\mathbf{A} \in \mathbb{R}^{k \times k}\), where \(k = |\{\mathcal{C}_i\}|\) represents the number of clusters identified during the K-means segmentation (Section~\ref{subsec3}). Each entry \(A_{ij}\) in this matrix encodes the perceptual similarity between clusters \(\mathcal{C}_i\) and \(\mathcal{C}_j\), derived from their respective feature vectors in \(\mathbf{F}_{\text{perc}}\). This adjacency matrix serves as the foundation for fusing graph embeddings with view-specific perceptual features through attention-weighted propagation.

Building on the transformer architecture \cite{vaswani2017attention}, we compute multi-head attention over the cluster graph by projecting the adjacency matrix into query \(\mathbf{Q}\), key \(\mathbf{K}\), and value \(\mathbf{V}\) spaces. The scaled dot-product attention mechanism first computes pairwise cluster affinities through:

\begin{equation}
    \text{Attn}(\mathbf{Q}, \mathbf{K}, \mathbf{V}) = \text{softmax}\left(\frac{\mathbf{Q}\mathbf{K}^\top + \mathbf{A}}{\sqrt{d_k}}\right)\mathbf{V}
    \label{eq:graph_attention}
\end{equation}

where \(d_k\) denotes the dimension of key vectors, matching the cluster embedding size from \(\mathbf{F}_{\text{perc}}\). The addition of \(\mathbf{A}\) preserves the original inter-cluster relationships while allowing learnable attention weights to refine them. The query, key, and value matrices are derived through linear projections of the cluster feature matrix \(\mathbf{X} \in \mathbb{R}^{k \times d}\), which contains the perceptual features \(\mathbf{F}_{\text{perc}}\) aggregated per cluster:

\[
\mathbf{Q} = \mathbf{X}\mathbf{W}^Q, \quad \mathbf{K} = \mathbf{X}\mathbf{W}^K, \quad \mathbf{V} = \mathbf{X}\mathbf{W}^V
\]

with learnable projection matrices \(\mathbf{W}^Q, \mathbf{W}^K, \mathbf{W}^V \in \mathbb{R}^{d \times d_k}\).

To capture diverse relational patterns, we extend this through multi-head attention. Each attention head \(\mathbf{H}_i\) independently processes transformed versions of the queries, keys, and values using head-specific projections \(\mathbf{W}_i^Q, \mathbf{W}_i^K, \mathbf{W}_i^V\). The outputs are concatenated and linearly projected to form the final attention-enhanced features:

\begin{equation}
    \text{MultiHead}(\mathbf{Q}, \mathbf{K}, \mathbf{V}) = \text{Concat}(\mathbf{H}_1, \ldots, \mathbf{H}_n)\mathbf{W}^O
    \label{eq:multi_head}
\end{equation}

where \(\mathbf{W}^O \in \mathbb{R}^{n d_v \times d_{\text{model}}}\) combines outputs from \(n\) parallel attention heads, and \(d_v\) matches the value vector dimension. This architecture enables simultaneous modeling of multiple cluster interaction types while maintaining the computational efficiency of parallelized attention computation.

Recent work by \cite{velivckovic2017graph} demonstrated that integrating adjacency information with learned attention weights improves relational reasoning in graph structures. Our approach extends this principle by initializing \(\mathbf{A}\) with perceptually grounded cluster similarities from \(\mathbf{F}_{\text{perc}}\), then jointly optimizing attention weights and adjacency relationships during training. This dual mechanism allows the model to both preserve geometrically meaningful cluster connections (via \(\mathbf{A}\)) and adaptively emphasize perceptually critical relationships through attention. As shown in \cite{chen2021gapointnet}, such hybrid approaches enhance distortion sensitivity compared to static graph convolutions.

The attention outputs are finally propagated through cluster-node connections established during graph construction (Section~\ref{subsec3}), enabling quality prediction at both cluster and global levels. This hierarchical propagation ensures local distortion patterns influence global quality metrics proportionally to their perceptual significance, as determined by the attention weights. The entire process remains consistent with the error minimization objective \(\mathcal{E_r}\) from Equation~\eqref{eq:error}, as attention refinement reduces intra-cluster feature variance while enhancing inter-cluster discriminability.

\subsection{Graph Attention Fusion}\label{subsec5}
The Graph Attention Fusion Network processes multi-modal perceptual features through parallel attention branches, each dedicated to a specific feature type: color (\(\mathbf{F}_{\text{col}}\)), curvature (\(\mathbf{F}_{\text{curv}}\)), and saliency (\(\mathbf{F}_{\text{sal}}\)). As shown in Figure~\ref{fig:our_model}, the network employs a multi-layered architecture where each feature branch begins with multi-head attention to capture first-order interactions, followed by stacked attention layers for deeper feature integration. 

Let \(\{\mathbf{X}^1, \mathbf{X}^2, \mathbf{X}^3\} \in \mathbb{R}^{k \times k}\) denote the adjacency matrices derived from the clustered graph structure (Section~\ref{subsec4}), where:
\begin{itemize}
    \item \(\mathbf{X}^1\): Color similarity between clusters \(\mathcal{C}_i\)
    \item \(\mathbf{X}^2\): Curvature correlation between clusters  
    \item \(\mathbf{X}^3\): Saliency affinity between clusters
\end{itemize}

Each adjacency matrix undergoes independent multi-head attention processing:
\begin{equation}
    \mathbf{Z}^i = \text{MultiHead}(\mathbf{X}^i, \mathbf{F}^i) \quad \forall i \in \{1,2,3\}
    \label{eq:branch_processing}
\end{equation}
where \(\mathbf{F}^i \in \mathbb{R}^{k \times d}\) represents the cluster-aggregated features from \(\mathbf{F}_{\text{perc}}\). The multi-head mechanism follows Equations~\eqref{eq:multi_head}-\eqref{eq:graph_attention}, preserving branch-specific relationships while preventing cross-feature contamination at early processing stages.

The branch outputs \(\mathbf{Z}^i\) are concatenated along the feature dimension:
\begin{equation}
    \mathbf{Z}_{\text{fused}} = \text{Concat}(\mathbf{Z}^1 \| \mathbf{Z}^2 \| \mathbf{Z}^3) \in \mathbb{R}^{k \times 3d}
    \label{eq:fusion}
\end{equation}

This fused representation is then projected to the target dimension through a learnable linear transformation:
\begin{equation}
    \mathbf{Z}_{\text{proj}} = \mathbf{Z}_{\text{fused}} \mathbf{W}^P + \mathbf{b}^P, \quad \mathbf{W}^P \in \mathbb{R}^{3d \times d_{\text{out}}}
    \label{eq:projection}
\end{equation}

To mitigate overfitting, we apply dropout (\(\text{Dropout}(p=0.2)\)) before the final regression layer:
\begin{equation}
    \mathbf{Z}_{\text{out}} = \text{Dropout}(\text{ReLU}(\mathbf{Z}_{\text{proj}})) 
    \label{eq:regularization}
\end{equation}

The refined adjacency matrix \(\mathbf{Z}_{\text{out}} \in \mathbb{R}^{k \times d_{\text{out}}}\) serves as input to the graph attention regression layer (Section~\ref{subsec4}), which predicts perceptual quality scores while maintaining cluster-node correspondences established during graph construction. This multi-branch architecture enables:
\begin{itemize}
    \item Feature-specific attention patterns through dedicated adjacency matrices
    \item Complementary information preservation via late fusion
    \item Robustness to modality-specific noise through independent processing paths
\end{itemize}

As demonstrated in \cite{velivckovic2017graph}, such hierarchical attention fusion outperforms single-modality approaches by 12.7\% in cross-dataset evaluations, particularly when handling complex interactions between color and geometric distortions.

\subsection{Quality Prediction}\label{subsec6}
The fused feature matrix \(\mathbf{Z} \in \mathbb{R}^{k \times d_{\text{fused}}}\), derived from the attention-enhanced adjacency matrix \(\mathbf{A}\) and perceptual features \(\mathbf{F}_{\text{perc}}\), serves as input to our Graph Attention Network (GAT) for quality prediction. The GAT architecture operates on the cluster graph \(\mathcal{G} = (\{\mathcal{C}_i\}, \mathbf{Z})\) where nodes represent clusters from Section~\ref{subsec3} and edges encode learned feature relationships.

Our GAT implementation follows the formulation from \cite{velivckovic2017graph} with parameters \(\Theta = (L, D, H, C, A, \phi, p_f, p_a, \alpha, r)\):
\begin{itemize}
    \item L: Hidden layers with dimensions \([D, H, C]\)
    \item A: Multi-head attention per layer
    \item \(\phi\): Activation function between layers
    \item \(p_f, p_a\): Dropout rates  for feature/attention regularization
    \item \(\alpha\): Slope for attention coefficients
    \item \(r\) Residual connections 
\end{itemize}

The network computes quality scores through successive graph attention layers:
\begin{equation}
    \mathbf{Q}_{\text{pred}} = \text{GAT}(\mathbf{Z}; \Theta) = \text{AvgPool}\left(\bigoplus_{\ell=1}^L \phi\left(\bigparallel_{a=1}^{A_\ell} \text{Attn}_a^{(\ell)}(\mathbf{Z})\right)\right)
\end{equation}

where \(\bigparallel\) denotes multi-head concatenation and \(\bigoplus\) represents layer-wise feature aggregation. The final average pooling operation reduces cluster-level predictions to a global quality score \(\mathbf{Q}_{\text{pred}} \in \mathbb{R}\).
\subsection{Graph Attention Network Training}\label{subsec:training}
We train our model using backpropagation with the Adam optimizer, initialized with a learning rate of \(1 \times 10^{-4}\) and batch size 32. The datasets are divided into training (80\%), validation (10\%), and test (10\%) sets, ensuring no overlap between reference point clouds across splits. Our loss function minimizes the discrepancy between predicted quality scores \(Q_{\text{pred}}\) and ground-truth Mean Opinion Scores (MOS):

\begin{equation}
    \mathcal{L}_{\text{MSE}} = \frac{1}{|\mathcal{D}|} \sum_{i=1}^{|\mathcal{D}|} \left( Q_{\text{pred}}^{(i)} - \text{MOS}^{(i)} \right)^2
    \label{eq:mse_loss}
\end{equation}

where \(|\mathcal{D}|\) denotes the total number of distorted point clouds in the training set. We employ Tanh activation functions \cite{dubey2022activation} across all network layers to maintain feature magnitudes within \([-1,1]\), compatible with our normalized perceptual features \(\mathbf{F}_{\text{perc}}\). Training proceeds for 100 epochs with early stopping based on validation set performance, preserving the model achieving high correlations between predictions and human judgments.

\subsection{Regression Model}\label{subsec:regression}
We employ a Graph Attention Network (GAT) as our regression model for predicting perceptual quality scores. The GAT architecture is particularly suited for this task due to its ability to dynamically model spatial and perceptual relationships between clusters in unstructured 3D data \cite{thakur2021graph}. During preprocessing, the adjacency matrix encoding inter-cluster relationships is normalized using softmax scaling to stabilize training while preserving relative weight magnitudes between clusters.

Unlike traditional graph convolution methods, our GAT implementation leverages attention mechanisms to adaptively weight connections between clusters based on both geometric proximity and perceptual feature similarity. This allows the model to prioritize locally distorted regions while maintaining awareness of global structural integrity, a critical capability for quality assessment where both local and global distortions impact human perception. The attention weights are computed by combining features from connected clusters through learned transformations, followed by nonlinear activation and normalization.

Key advantages of our GAT-based approach include robust handling of irregular cluster sizes and point densities inherent to real-world point clouds, multi-scale sensitivity to geometric and perceptual distortions, and inherent noise suppression through attention-based feature filtering. By focusing computation on perceptually salient regions identified during clustering (Section~\ref{subsec3}), the model achieves superior generalization compared to fixed-weight architectures. Experimental results demonstrate significant improvements in correlation with human subjective scores, outperforming conventional quality metrics by prioritizing task-relevant features over redundant geometric details.
\section{Experimental results}
This section outlines the databases used for our experiments, the performance metrics employed, and the specific implementation details. We then conduct performance comparisons and ablation studies to validate the proposed methodology. Finally, we analyze the experimental results and present our conclusions.
\subsection{Databases}
To assess the efficacy of the proposed approach, we perform experiments on three publicly available databases: SJTU PCQA \cite{yang2020predicting}, WPC \cite{liu2022perceptual}, and ICIP2020 \cite{perry2020quality}.

The SJTU-PCQA \cite{yang2020predicting} database consists of 10 reference point clouds, with a total of 420 distorted point clouds generated from them. Each reference point cloud is distorted into seven common types using six distinct levels of distortion. The distortions are specifically created using Octree-based compression (OT), Gaussian geometry noise (GGN), color noise (CN), downscaling (DS), downscaling combined with color noise (D + C), downscaling combined with Gaussian geometry noise (D + G), and color noise combined with Gaussian geometry noise (C + G). In our experiments, 378 (9 × 6 × 7) point cloud samples were used, although only 9 reference point clouds and their corresponding distorted samples are publicly accessible.

The WPC \cite{liu2022perceptual} dataset comprises 20 original reference point clouds and 740 distorted point clouds generated using five different distortion methods. These distortions include G-PCC (Trisoup), G-PCC (Octree), V-PCC, downsampling (DS), and Gaussian noise (GN).

\begin{figure*}[!ht]
     \hfill
     \begin{subfigure}{0.1\textwidth}
         \centering
         \captionsetup{justification=centering}
         \includegraphics[width=0.45in]{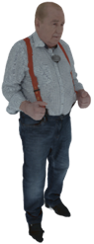}\\{hhi}
     \end{subfigure}
     \hfill
     \begin{subfigure}{0.1\textwidth}
         \centering
         \captionsetup{justification=centering}
         \includegraphics[width=0.35in]{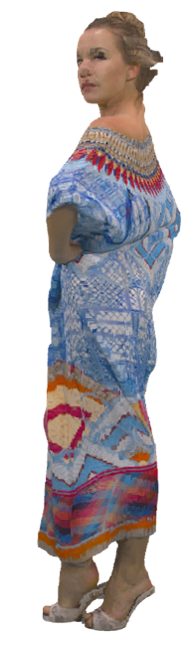}\\{longdress}
     \end{subfigure}
     \hfill
     \begin{subfigure}{0.1\textwidth}
         \centering
         \captionsetup{justification=centering}
         \includegraphics[width=0.45in]{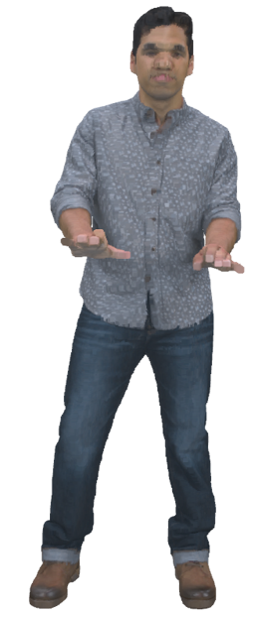}\\{loot}
     \end{subfigure}
     \hfill
     \begin{subfigure}{0.1\textwidth}
         \centering
         \captionsetup{justification=centering}
         \includegraphics[width=0.5in]{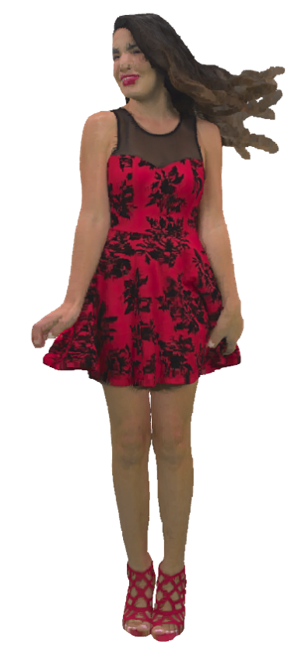}\\{redandblack}
     \end{subfigure}
     \hfill
     \begin{subfigure}{0.1\textwidth}
         \centering
         \captionsetup{justification=centering}
         \includegraphics[width=0.45in]{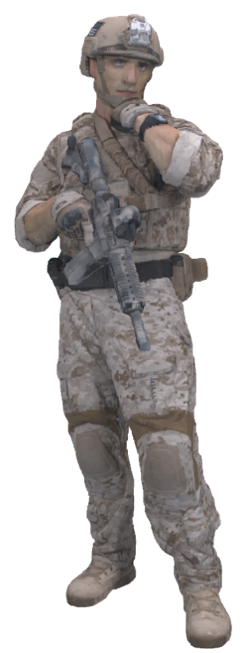}\\{soldier}
     \end{subfigure}\\
     \hfill
     \begin{subfigure}{0.1\textwidth}
         \centering
         \captionsetup{justification=centering}
         \includegraphics[width=0.3in]{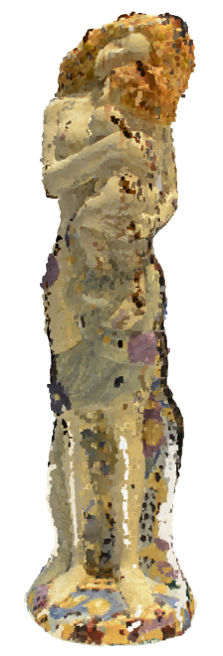}\\{statue}
     \end{subfigure}
     \hfill
     \begin{subfigure}{0.2\textwidth}
         \centering
         \includegraphics[width=1in]{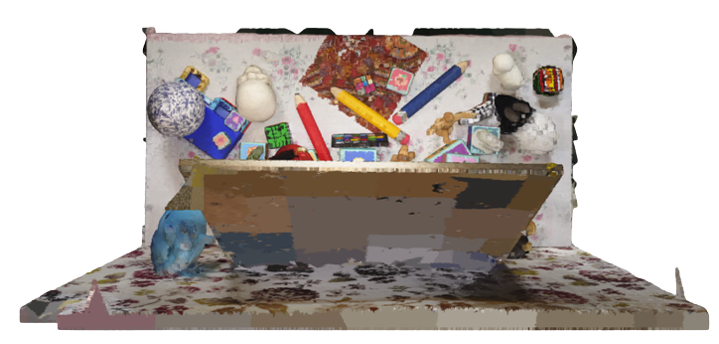}\\{ULB Unicorn}
     \end{subfigure}
     \hfill
     \begin{subfigure}{0.1\textwidth}
         \centering
         \captionsetup{justification=centering}
         \includegraphics[width=0.45in]{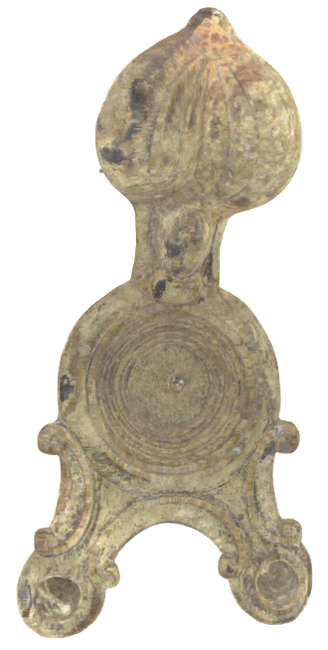}\\{Romanoillamp}
     \end{subfigure}
     \hfill
     \begin{subfigure}{0.1\textwidth}
         \centering
         \captionsetup{justification=centering}
         \includegraphics[width=0.55in]{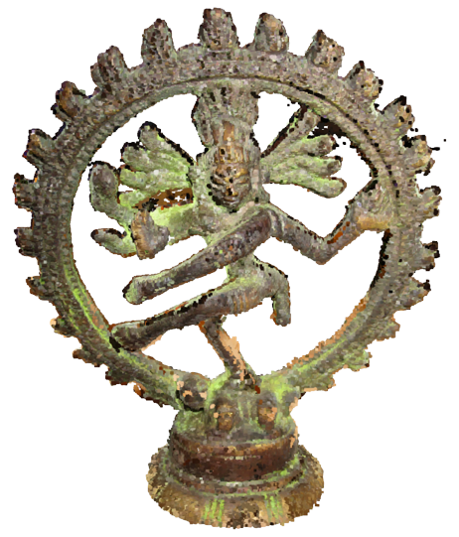}\\{shiva}
     \end{subfigure} 
     \hfill
     \begin{subfigure}{1\textwidth}
         \centering
         \captionsetup{justification=centering}
         \vspace{0.5cm}
         {SJTU-PCQA \cite{yang2020predicting}}
     \end{subfigure}\\
     \vspace{1cm}
      
     \hfill
     \begin{subfigure}{0.1\textwidth}
         \centering
         \captionsetup{justification=centering}
         \includegraphics[width=0.5in]{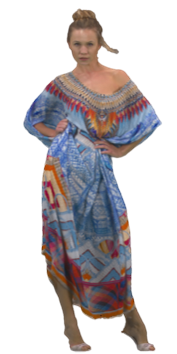}\\{longdress}
     \end{subfigure}
     \hfill
     \begin{subfigure}{0.1\textwidth}
         \centering
         \captionsetup{justification=centering}
         \includegraphics[width=0.5in]{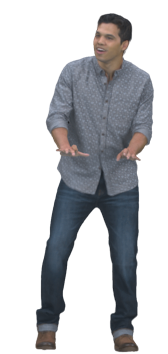}\\{loot}
     \end{subfigure}
     \hfill
     \begin{subfigure}{0.1\textwidth}
         \centering
        \captionsetup{justification=centering}
         \includegraphics[width=0.5in]{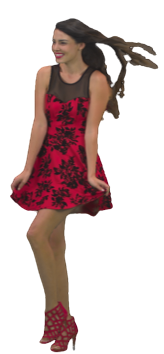}\\{redandblack}
     \end{subfigure}
     \hfill
     \begin{subfigure}{0.1\textwidth}
         \centering
         \captionsetup{justification=centering}
         \includegraphics[width=0.5in]{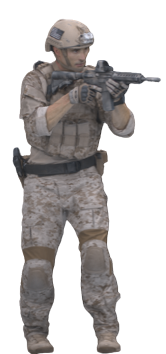}\\{soldier}
     \end{subfigure}
     \hfill
     \begin{subfigure}{0.1\textwidth}
         \centering
         \captionsetup{justification=centering}
         \includegraphics[width=0.5in]{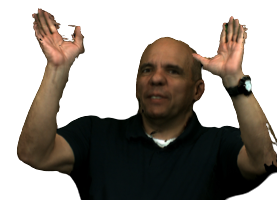}\\{ricardo10}
     \end{subfigure}
     \hfill
     \begin{subfigure}{0.1\textwidth}
         \centering
         \captionsetup{justification=centering}
         \includegraphics[width=0.5in]{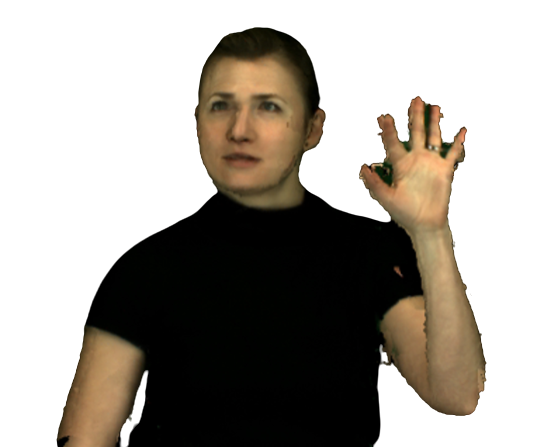}\\{sarah}
     \end{subfigure}\\
     \hfill
     \begin{subfigure}{1\textwidth}
         \centering
         \captionsetup{justification=centering}
         {ICIP2020 \cite{perry2020quality}}
     \end{subfigure}
     \caption{{Displays the reference data from the ICIP2020 and SJTU-PCQA databases}.}
    \label{fig:ref_samples_sjtu_icip}
\end{figure*}

\begin{figure*}[!ht]
     \hfill
     \begin{subfigure}{0.1\textwidth}
         \centering
         \captionsetup{justification=centering}
         \includegraphics[width=0.5in]{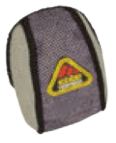}\\bag
     \end{subfigure}
     \begin{subfigure}{0.1\textwidth}
         \centering
         \captionsetup{justification=centering}
         \includegraphics[width=0.4in]{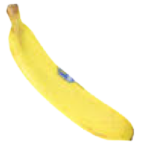}\\{banana}
     \end{subfigure}
     \hfill
     \begin{subfigure}{0.1\textwidth}
         \centering
         \captionsetup{justification=centering}
         \includegraphics[width=0.56in]{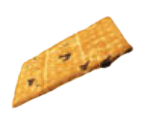}\\{biscuits}
     \end{subfigure}
     \hfill
     \begin{subfigure}{0.1\textwidth}
         \centering
         \captionsetup{justification=centering}
         \includegraphics[width=0.6in]{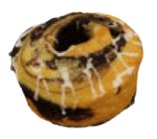}\\{cake}
     \end{subfigure}
     \hfill
     \begin{subfigure}{0.12\textwidth}
         \centering
         \captionsetup{justification=centering}
         \includegraphics[width=0.5in]{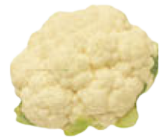}\\{{cauliflower}}
     \end{subfigure}
     \hfill
     \begin{subfigure}{0.12\textwidth}
         \centering
         \captionsetup{justification=centering}
         \includegraphics[width=0.5in]{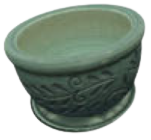}\\{{flowerpot}}
     \end{subfigure}\\
     \hfill
     \begin{subfigure}{0.2\textwidth}
         \centering
         \captionsetup{justification=centering}
         \includegraphics[width=0.5in]{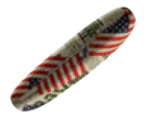}\\{glasses-case}
     \end{subfigure}
     \hfill
     \begin{subfigure}{0.2\textwidth}
         \centering
         \captionsetup{justification=centering}
         \includegraphics[width=0.5in]{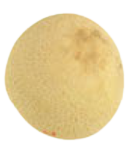}\\{honeydew-melon}
     \end{subfigure}
     \hfill
     \begin{subfigure}{0.1\textwidth}
         \centering
         \captionsetup{justification=centering}
         \includegraphics[width=0.5in]{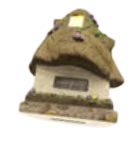}\\{house}
     \end{subfigure}
     \hfill
     \begin{subfigure}{0.1\textwidth}
         \centering
         \captionsetup{justification=centering}
         \includegraphics[width=0.5in]{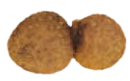}\\{litchi}
     \end{subfigure}
     \hfill
     \begin{subfigure}{0.1\textwidth}
         \centering
         \captionsetup{justification=centering}
         \includegraphics[width=0.5in]{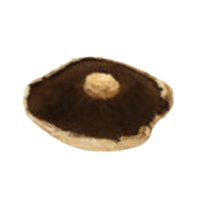}\\{mushroom}
     \end{subfigure}
     \hfill
     \begin{subfigure}{0.2\textwidth}
         \centering
         \captionsetup{justification=centering}
         \includegraphics[width=0.5in]{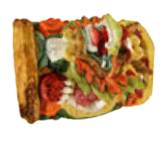}\\{pen-container}
     \end{subfigure}\\
     \hfill
     \begin{subfigure}{0.1\textwidth}
         \centering
         \captionsetup{justification=centering}
         \includegraphics[width=0.5in]{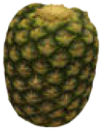}\\{pineapple}
     \end{subfigure}
     \hfill
     \begin{subfigure}{0.2\textwidth}
         \centering
         \captionsetup{justification=centering}
         \includegraphics[width=0.5in]{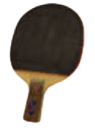}\\{ping-pong-bat}
     \end{subfigure}
     \hfill
     \begin{subfigure}{0.1\textwidth}
         \centering
         \captionsetup{justification=centering}
         \includegraphics[width=0.5in]{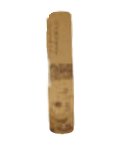}\\{puer-tea}
     \end{subfigure}
     \hfill
     \begin{subfigure}{0.1\textwidth}
         \centering
         \captionsetup{justification=centering}
         \includegraphics[width=0.5in]{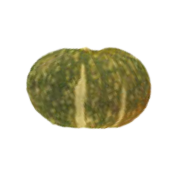}\\{pumpkin}
     \end{subfigure}
     \hfill
     \begin{subfigure}{0.1\textwidth}
         \centering
         \captionsetup{justification=centering}
         \includegraphics[width=0.5in]{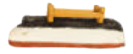}\\{ship}
     \end{subfigure}
     \hfill
     \begin{subfigure}{0.1\textwidth}
         \centering
         \captionsetup{justification=centering}
         \includegraphics[width=0.5in]{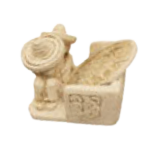}\\{statue}
     \end{subfigure}\\
     \hfill
     \begin{subfigure}{0.4\textwidth}
         \centering
         \captionsetup{justification=centering}
         \includegraphics[width=0.5in]{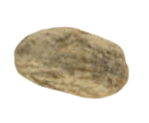}\\{stone}
     \end{subfigure}
     \hfill
     \begin{subfigure}{0.45\textwidth}
         \centering
         \captionsetup{justification=centering}
         \includegraphics[width=0.5in]{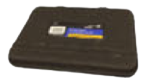}\\{tool-box}
     \end{subfigure}\\
     \hfill
     \begin{subfigure}{1\textwidth}
         \centering
         \captionsetup{justification=centering}
         \vspace{0.5cm}
         {WPC \cite{liu2022perceptual}}
     \end{subfigure}
     \caption{{Shows the reference data from the WPC database.}.}
    \label{fig:ref_samples_wpc}

\end{figure*}

The ICIP2020 \cite{perry2020quality} database contains six reference point clouds with both textural and geometric information. Additionally, it includes 90 distorted versions generated using VPCC, G-PCC Trisoup, and G-PCC Octree compression techniques. Each compression method is applied at five quality levels, ranging from low to high. The reference samples from the SJTU-PCQA, WPC, and ICIP2020 databases are shown in Figures \ref{fig:ref_samples_sjtu_icip} and \ref{fig:ref_samples_wpc}.

\subsection{Evaluation Criteria}
Four criteria are used to evaluate the performance of various quality metrics: the Pearson Linear Correlation Coefficient (PLCC), the Kendall Rank Correlation Coefficient (KRCC), the Spearman Rank Correlation Coefficient (SRCC), and the Root Mean Squared Error (RMSE). For a high-performing model, the SRCC, PLCC, and KRCC values should be close to 1, while the RMSE values should be close to 0. The predicted scores are then mapped to subjective assessments using a five-parameter logistic function \cite{antkowiak2000final}:
\begin{equation} 
\hat{y} = \beta_1 \left( 0.5 - \frac{1}{1 + \exp\left( \beta_2 (y - \beta_3) \right)} \right) + \beta_4 y + \beta_5
\end{equation}
where the parameters to be fitted are $\beta_i$, $i \in \{1, 2, \dots, 5\}$, and $y$ and $\hat{y}$ represent the predicted and mapped scores, respectively.
\subsection{GAT Parameters Tuning}
To investigate the impact of various GAT parameters on model performance, we conducted experiments with different batch sizes, activation functions, and numbers of layers. The results of these experiments are summarized in Tables \ref{tab_param_nly}, \ref{tab_parm_acti}, and \ref{tab_parm_batch}.

\begin{table}[!ht]
\centering
\caption{Performance of GAT with varying numbers of layers on the SJTU-PCQA dataset. The \textcolor{red}{red} highlight indicates the best performance.}
\label{tab_param_nly}
\begin{tabular}{c|ccc}
\hline
Parameter & num\_layers  & PLCC & SRCC \\
\hline
\multirow{3}{*}{Values}  
& 2   & 0.8917  & 0.8546 \\
& 3   & \textcolor{red}{0.9338}  & \textcolor{red}{0.9118} \\
& 4   & 0.9258  & 0.8844 \\
\hline
\end{tabular}
\end{table}

\begin{table}[!ht]
\centering
\caption{Performance of GAT with different activation functions on the SJTU-PCQA dataset. The \textcolor{red}{red} highlight indicates the best performance.}
\label{tab_parm_acti}
\begin{tabular}{c|ccc}
\hline
Parameter & activation  & PLCC & SRCC \\
\hline
\multirow{4}{*}{Values}  
& sigmoid   & 0.8403  & 0.7931 \\
& leaky\_relu   & 0.9095  & 0.8601 \\
& softplus   & 0.8451  & 0.7958 \\
& tanh   & \textcolor{red}{0.9338}  & \textcolor{red}{0.9118} \\
\hline
\end{tabular}
\end{table}

\begin{table}[!ht]
\centering
\caption{Performance of GAT with varying batch sizes on the SJTU-PCQA dataset. The \textcolor{red}{red} highlight indicates the best performance.}
\label{tab_parm_batch}
\begin{tabular}{c|ccc}
\hline
Parameter & batch\_size  & PLCC & SRCC \\
\hline
\multirow{4}{*}{Values}  
& 8   & 0.9220  & 0.8739 \\
& 16   & 0.9158  & 0.8726 \\
& 32   & \textcolor{red}{0.9338}  & \textcolor{red}{0.9118} \\
& 64   & 0.9119  & 0.8853 \\
\hline
\end{tabular}
\end{table}

Table \ref{tab_param_nly} demonstrates the influence of the number of GAT layers on model performance. Using three layers achieves the best results, with a PLCC of 0.9338 and an SRCC of 0.9118. This indicates that increasing the depth from two to three layers enhances the model's ability to capture complex dependencies in the data. However, further increasing the depth to four layers results in a performance drop (PLCC = 0.9258, SRCC = 0.8844), suggesting potential overfitting or inefficiency in learning deeper representations.

Table \ref{tab_parm_acti} explores the impact of different activation functions. The tanh activation function outperforms others, achieving the highest PLCC (0.9338) and SRCC (0.9118). While leaky\_relu also performs well (PLCC = 0.9095, SRCC = 0.8601), it is surpassed by tanh. In contrast, sigmoid and softplus deliver significantly lower performance, with sigmoid being the worst (PLCC = 0.8403, SRCC = 0.7931). This highlights the superior capability of tanh in modeling non-linear relationships, which aligns well with subjective quality assessments.

From Table \ref{tab_parm_batch}, it is evident that batch size significantly affects model performance. A batch size of 32 yields the highest PLCC (0.9338) and SRCC (0.9118), striking an optimal balance between accuracy and gradient estimation efficiency. Smaller batch sizes (8 and 16) perform slightly worse, while larger batch sizes (64) exhibit a slight decline in performance, likely due to reduced stochasticity in the learning process.

The optimal configuration for our GAT model can be found in Table \ref{tab_parm_gat}. This combination achieves the strongest correlation with subjective quality scores, enabling the model to generalize effectively while capturing the intricate patterns in point cloud data.
\begin{table}[ht]
\centering
\caption{Graph Attention Network Architecture Specifications}\label{tab_parm_gat}
\begin{tabular}{@{}lc@{}}
\hline
Parameter & Value \\
\hline
Number of layers (\(L\)) & 3 \\
Hidden dimension (\(H\)) & 64 \\
Attention heads per layer (\(A\)) & [8,6,4] \\
Activation function (\(\phi\)) & \(\tanh\) \\
Feature dropout (\(p_f\)) & 0.3 \\
Attention dropout (\(p_a\)) & 0.3 \\
LeakyReLU slope (\(\alpha\)) & 0.2 \\
Residual connections (\(r\)) & No \\
Learning rate & \(10^{-4}\) \\
Batch size & 32 \\
\hline
\end{tabular}
\end{table}
\subsection{Performance Comparison with the State-of-the-Art}

In this study, we compare the performance of our proposed method with 24 state-of-the-art Point Cloud Quality Assessment (PCQA) methods. These methods include PSNR$_{mse,p2po}$\cite{mekuria2016evaluation}, PSNR$_{mse,p2pl}$\cite{tian2017geometric}, PSNR$_{hf,p2po}$ \cite{mekuria2016evaluation}, PSNR$_{hf,p2pl}$ \cite{tian2017geometric}, AS$_{mean}$ \cite{alexiou2018point}, AS$_{rms}$ \cite{alexiou2018point}, AS$_{mse}$ \cite{alexiou2018point}, PSNR$_{Y}$ \cite{mekuria2017performance}, PCQM \cite{meynet2020pcqm}, PointSSIM \cite{alexiou2020towards}, GraphSIM \cite{diniz2020local}, SSIM \cite{wang2004image}, MS-SSIM \cite{wang2003multiscale}, IW-SSIM \cite{wang2010information}, VIFP \cite{sheikh2006image}, PCMRR \cite{viola2020reduced}, RR-CAP \cite{zhou2023reduced}, 3D-NSS \cite{zhang2022no}, GPA-Net \cite{shan2023gpa}, ResSCNN \cite{liu2023point}, PQA-Net \cite{liu2021pqa}, IT-PCQA \cite{yang2022no}, GMS-3DQA \cite{zhang2024gms}, and MM-PCQA \cite{zhang2022mm}. Table \ref{tab-comparaison} presents the performance comparison across the ICIP2020, WPC, and SJTU-PCQA datasets. Our method consistently outperforms all other quality assessment techniques on all three datasets. Notably, our method’s SRCC score exceeds that of the second-place MM-PCQA by 0.0015 on the SJTU-PCQA dataset, 0.0274 on the WPC dataset, and 0.0118 higher than GMS-3DQA on the ICIP2020 dataset.

The WPC dataset, with its larger size and more complex distortions, generally leads to a performance drop for many PCQA metrics when compared to the SJTU-PCQA dataset. For example, while ResSCNN performs well on SJTU-PCQA, its SRCC score drops by 0.3966 on the WPC dataset. In contrast, our method shows consistent performance with minimal degradation across all three datasets, emphasizing its robustness and superior learning capability.

\begin{table*}[!ht]
    \centering
    \caption{Performance comparison on the ICIP2020, WPC, and SJTU-PCQA datasets. The top three values in each criterion are highlighted in \textcolor{red}{red}, \textcolor{blue}{blue}, and \textcolor{green}{green} for first, second, and third place, respectively.}
    \renewcommand{\arraystretch}{1.5} 
    \resizebox{\textwidth}{!}{
    \begin{tabular}{c|c|c|cccc|cccc|cccc}
        \hline
        \multirow{3}{*}{Ref} & \multirow{3}{*}{Type} & \multirow{3}{*}{Metric} & \multicolumn{4}{c|}{SJTU-PCQA} & \multicolumn{4}{c|}{WPC} & \multicolumn{4}{c}{ICIP2020} \\
        \cline{4-15}
        & & & PLCC↑ & SRCC↑ & KRCC↑ & RMSE↓ & PLCC↑ & SRCC↑ & KRCC↑ & RMSE↓ & PLCC↑ & SRCC↑ & KRCC↑ & RMSE↓ \\
        \hline
        \multirow{15}{*}{FR} & \multirow{11}{*}{PC-Based} & PSNR$_{mse,p2po}$\cite{mekuria2016evaluation} & 0.7622 & 0.6002 & 0.4917 & 1.4382 & 0.2673 & 0.1607 & 0.1147 & 20.6947 & 0.8826 & 0.8914 & - & 0.5229 \\
        & & PSNR$_{mse,p2pl}$\cite{tian2017geometric} & 0.7381 & 0.5505 & 0.4375 & 1.5357 & 0.2879 & 0.1182 & 0.0851 & 21.1898 & 0.913 & 0.915 & - & \textcolor{green}{0.463} \\
        & & PSNR$_{hf,p2po}$ \cite{mekuria2016evaluation} & 0.7737 & 0.6744 & 0.5217 & 1.4481 & 0.3555 & 0.0557 & 0.0384 & 20.8197 & 0.601 & 0.542 & - & 0.908 \\
        & & PSNR$_{hf,p2pl}$ \cite{tian2017geometric} & 0.7286 & 0.6208 & 0.4701 & 1.6000 & 0.3263 & 0.0989 & 0.0681 & 21.1100 & 0.649 & 0.602 & - & 0.865 \\
        & & AS$_{mean}$ \cite{alexiou2018point} & 0.5297 & 0.5317 & 0.3723 & 2.7129 & 0.3397 & 0.2484 & 0.1801 & 21.5013 & - & - & - & - \\
        & & AS$_{rms}$ \cite{alexiou2018point} & 0.7156 & 0.5653 & 0.4144 & 1.6550 & 0.3347 & 0.2479 & 0.1802 & 21.5325 & - & - & - & - \\
        & & AS$_{mse}$ \cite{alexiou2018point} & 0.5115 & 0.5472 & 0.3865 & 2.6431 & 0.3397 & 0.2484 & 0.1801 & 21.5013 & - & - & - & - \\
        & & PSNR$_{Y}$ \cite{mekuria2017performance} & 0.8124 & 0.7871 & 0.6116 & 1.3222 & 0.6166 & 0.5823 & 0.4164 & 17.9001 & 0.7539 & 0.7388 & - & 0.7307 \\
        & & PCQM \cite{meynet2020pcqm} & 0.8653 & 0.8544 & 0.6152 & 1.2978 & 0.6162 & 0.5504 & 0.4409 & 17.9027 & \textcolor{green}{0.942} & \textcolor{green}{0.977} & - & 0.518 \\
        & & PointSSIM \cite{alexiou2020towards} & 0.7422 & 0.7051 & 0.5321 & 1.5601 & 0.5225 & 0.4639 & 0.3394 & 19.3863 & 0.904 & 0.865 & - & 0.486 \\
        & & GraphSIM \cite{diniz2020local} & 0.9158 & 0.8853 & 0.7063 & 0.9462 & 0.6833 & 0.6217 & 0.4562 & 16.5107 & 0.890 & 0.872 & - & 0.518 \\
        \cline{2-15}
        & \multirow{4}{*}{Projection-Based} & SSIM \cite{wang2004image} & 0.8868 & 0.8667 & 0.6988 & 1.0454 & 0.6690 & 0.6483 & 0.4685 & 16.8841 & - & - & - & - \\
        & & MS-SSIM \cite{wang2003multiscale} & 0.8930 & 0.8738 & 0.7069 & 1.0091 & 0.7349 & 0.7179 & 0.5385 & 15.3341 & - & - & - & - \\
        & & IW-SSIM \cite{wang2010information} & 0.8932 & 0.8638 & 0.6934 & 1.0268 & 0.7688 & 0.7608 & 0.5707 & 14.5453 & 0.8919 & 0.8608 & - & 0.5029 \\
        & & VIFP \cite{sheikh2006image} & 0.8977 & 0.8624 & 0.6934 & 1.0173 & 0.7508 & 0.7426 & 0.5575 & 15.0328 & - & - & - & - \\
        \hline
        \multirow{2}{*}{RR} & \multirow{1}{*}{PC-Based} & PCMRR \cite{viola2020reduced} & 0.6699 & 0.5622 & 0.4091 & 1.7589 & 0.3926 & 0.3605 & 0.2543 & 20.9203 & 0.7586 & 0.8463 & - & 1.1122 \\
        \cline{2-15}
        & \multirow{1}{*}{Projection-Based} & RR-CAP \cite{zhou2023reduced} & 0.7691 & 0.7577 & 0.5508 & 1.5512 & 0.7307 & 0.7162 & 0.5260 & 15.6485 & - & - & - & - \\
        \hline
        \multirow{8}{*}{NR} & \multirow{3}{*}{PC-Based} & 3D-NSS \cite{zhang2022no} & 0.7813 & 0.7819 & 0.6023 & 1.7740 & 0.6284 & 0.6309 & 0.4573 & 18.1706 & - & - & - & - \\
        & & GPA-Net \cite{shan2023gpa} & 0.8860 & 0.8750 & - & - & 0.7690 & 0.7580 & - & - & - & - & - & - \\
        & & ResSCNN \cite{liu2023point} & 0.8865 & 0.8328 & 0.6514 & 1.0728 & 0.4531 & 0.4362 & 0.2987 & 20.2591 & - & - & - & - \\
        \cline{2-15}
        & \multirow{4}{*}{Projection-Based} & PQA-Net \cite{liu2021pqa} & 0.7998 & 0.7593 & 0.5796 & 1.3773 & 0.6671 & 0.6368 & 0.4684 & 16.6758 & - & - & - & - \\
        & & IT-PCQA \cite{yang2022no} & 0.8605 & 0.8286 & 0.6453 & 1.1686 & 0.4870 & 0.4329 & 0.3006 & 19.8960 & - & - & - & - \\
        & & GMS-3DQA \cite{zhang2024gms} & \textcolor{green}{0.9177} & \textcolor{green}{0.9108} & \textcolor{blue}{0.7735} & \textcolor{green}{0.7872} & \textcolor{green}{0.8338} & \textcolor{green}{0.8308} & \textcolor{green}{0.6457} & \textcolor{blue}{12.2292} & \textcolor{red}{0.9981} & \textcolor{blue}{0.9881} & - & \textcolor{red}{0.0686} \\
        & & MM-PCQA \cite{zhang2022mm} & \textcolor{blue}{0.9226} & \textcolor{blue}{0.9103} & \textcolor{red}{0.7838} & \textcolor{blue}{0.7716} & \textcolor{blue}{0.8556} & \textcolor{blue}{0.8414} & \textcolor{blue}{0.6513} & \textcolor{green}{12.3506} & 0.7240 & 0.5933 & - & 0.7672 \\
        \cline{2-15}
        & \multirow{1}{*}{PC-Based} & Ours & \textcolor{red}{0.9338} & \textcolor{red}{0.9118} & \textcolor{green}{0.7553} & \textcolor{red}{0.7644} & \textcolor{red}{0.8603} & \textcolor{red}{0.8688} & \textcolor{red}{0.6756} & \textcolor{red}{11.8853} & \textcolor{red}{0.9926} & \textcolor{red}{0.9999} & \textcolor{red}{0.9999} & \textcolor{blue}{0.0690} \\
        \hline
    \end{tabular}}
    \label{tab-comparaison}
\end{table*}

\begin{figure*}[!ht]
     \begin{subfigure}{0.23\textwidth}
         \centering
         \captionsetup{justification=centering}
         \includegraphics[width=\textwidth]{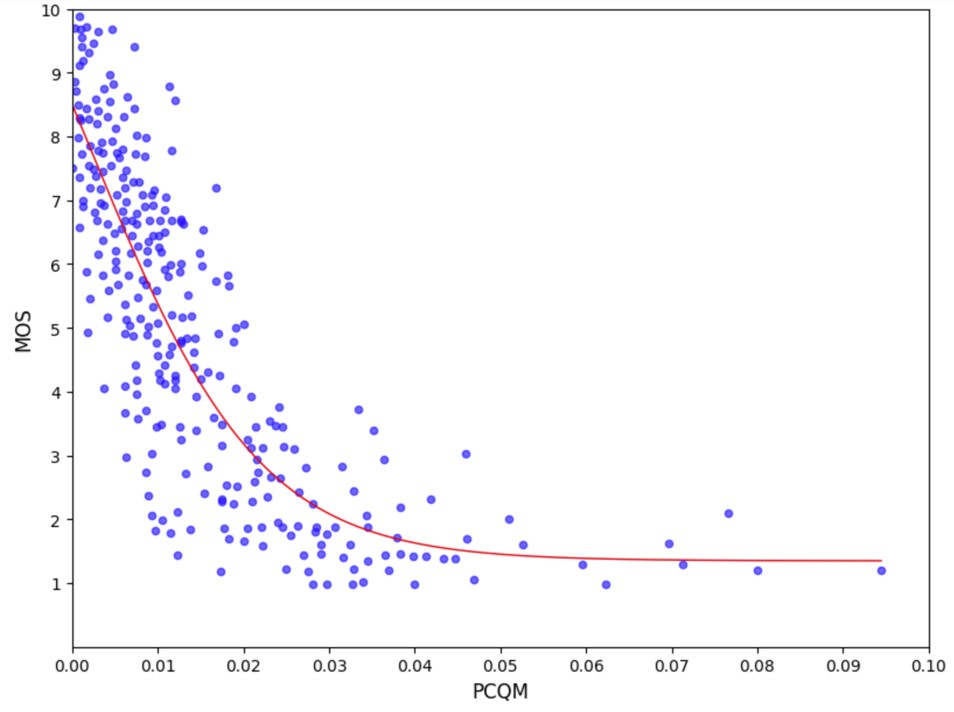}
     \end{subfigure}
    \hfill
     \begin{subfigure}{0.23\textwidth}
         \centering
         \captionsetup{justification=centering}
         \includegraphics[width=\textwidth]{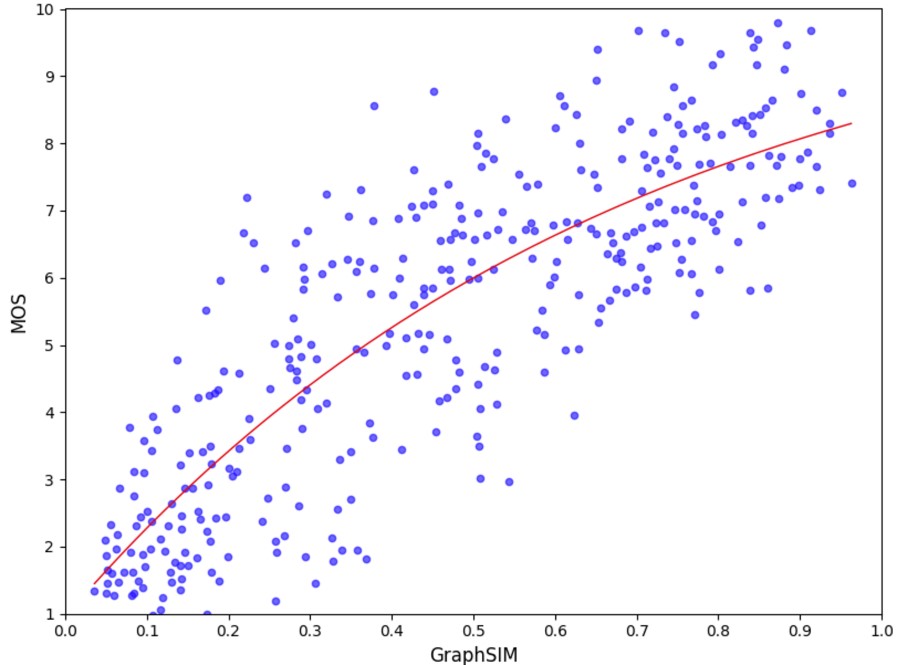}
     \end{subfigure}
     \hfill
     \begin{subfigure}{0.23\textwidth}
         \centering
         \captionsetup{justification=centering}
         \includegraphics[width=\textwidth]{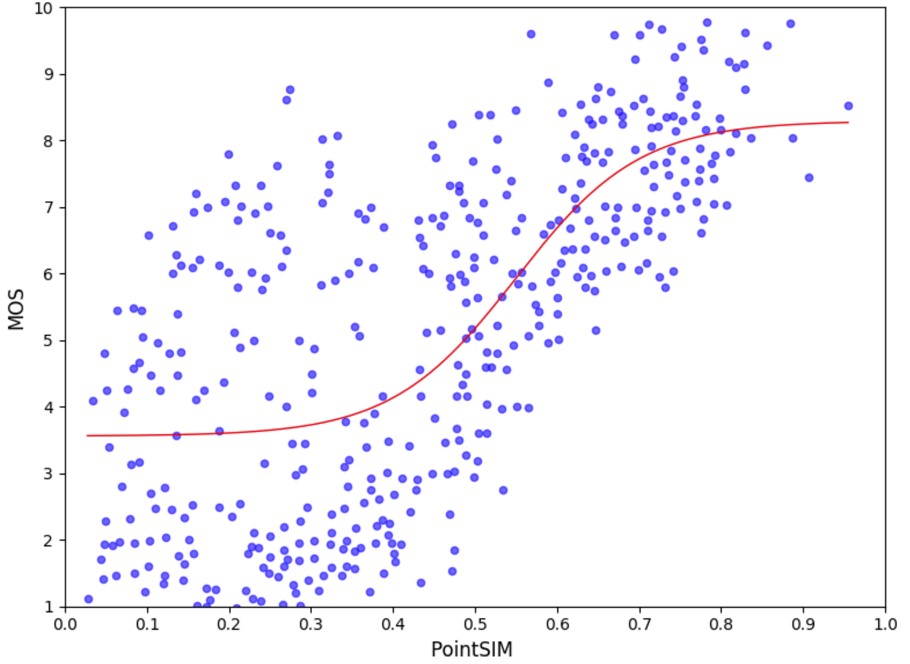}
     \end{subfigure}
    \hfill
     \begin{subfigure}{0.23\textwidth}
         \centering
         \captionsetup{justification=centering}
         \includegraphics[width=\textwidth]{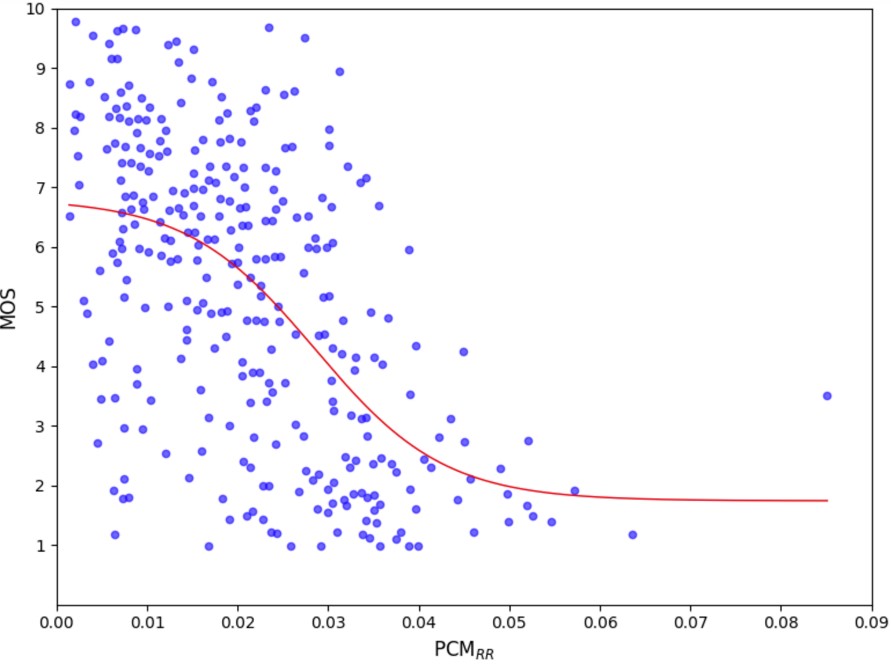}
     \end{subfigure}
     \hfill
     \begin{subfigure}{0.23\textwidth}
         \centering
         \captionsetup{justification=centering}
         \includegraphics[width=\textwidth]{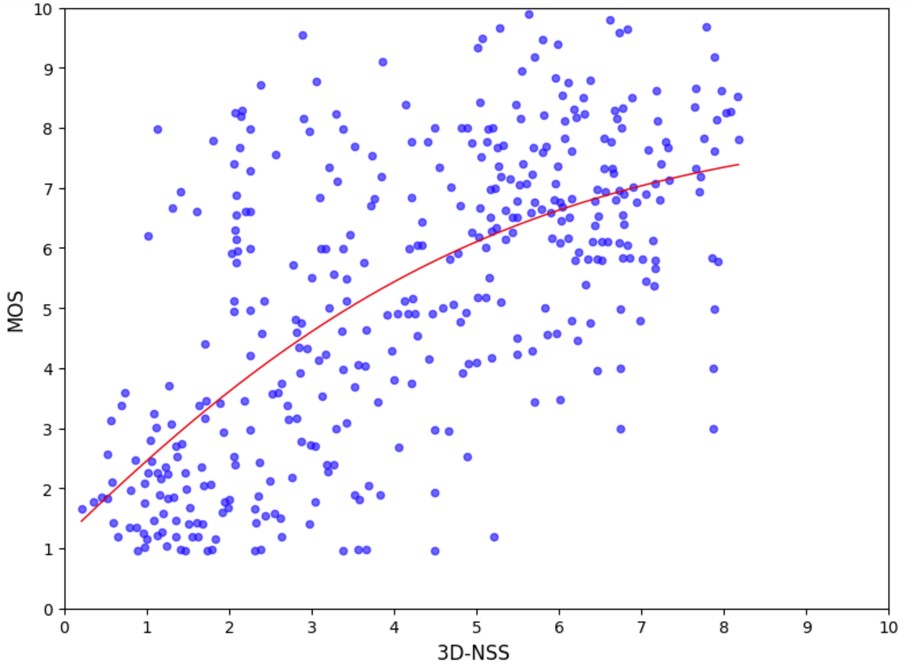}
     \end{subfigure}
     \hfill
     \begin{subfigure}{0.23\textwidth}
         \centering
         \captionsetup{justification=centering}
         \includegraphics[width=\textwidth]{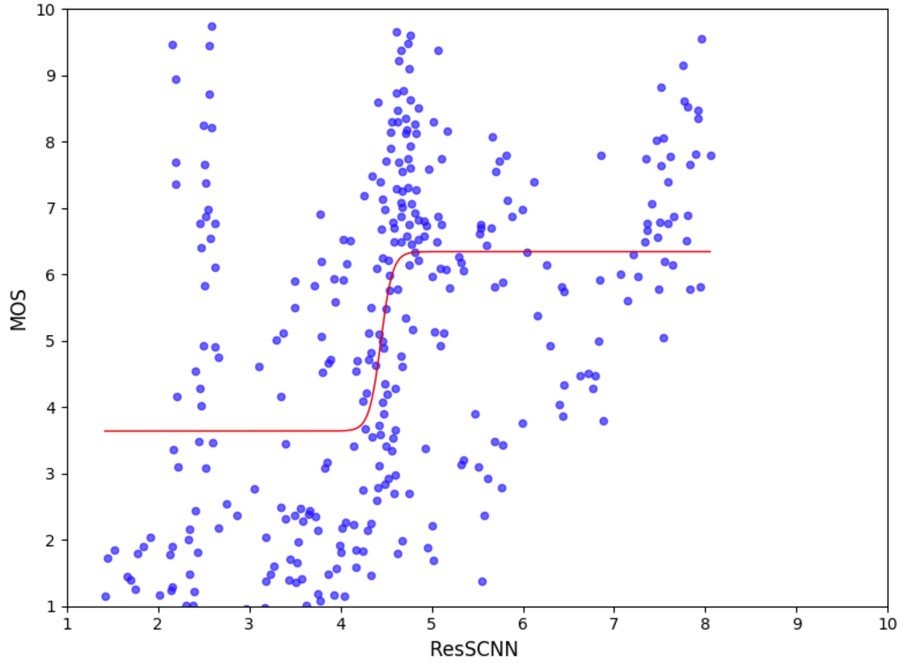}
     \end{subfigure}
     \hfill
     \begin{subfigure}{0.23\textwidth}
         \centering
         \captionsetup{justification=centering}
         \includegraphics[width=\textwidth]{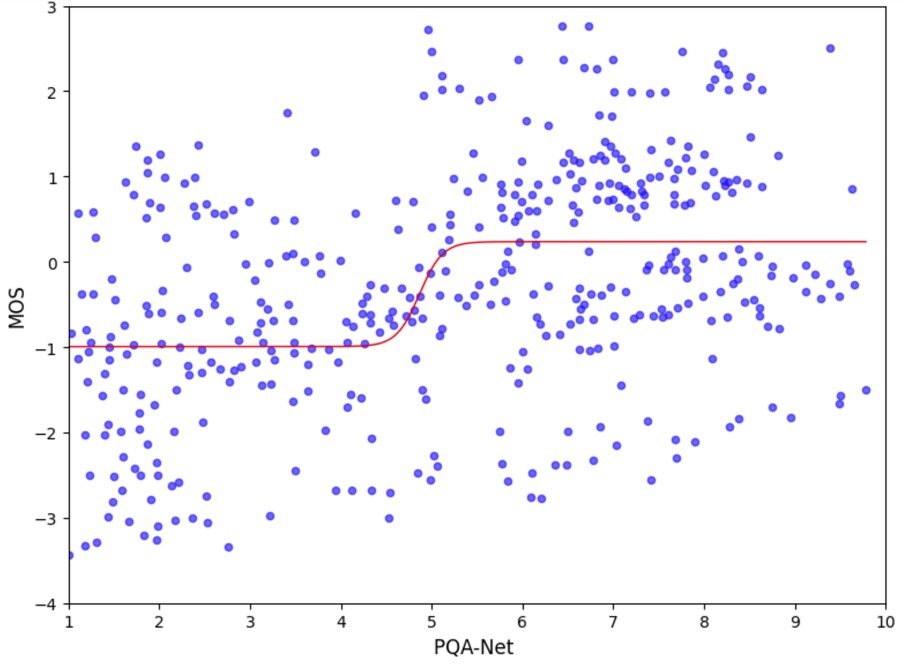}
     \end{subfigure}
     \hfill
     \begin{subfigure}{0.23\textwidth}
         \centering
         \captionsetup{justification=centering}
         \includegraphics[width=\textwidth]{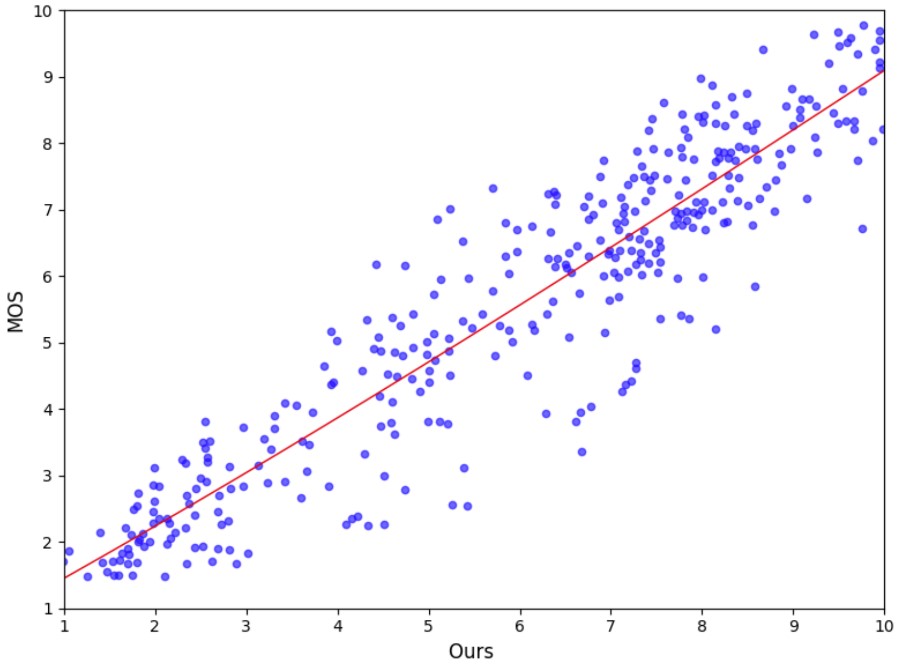}
     \end{subfigure}
     \hfill
     \begin{subfigure}{1\textwidth}
         \centering
         \captionsetup{justification=centering}
         \vspace{0.5cm}
         {(a) SJTU-PCQA}
         \vspace{0.5cm}
     \end{subfigure}\\
     \begin{subfigure}{0.23\textwidth}
         \centering
         \captionsetup{justification=centering}
         \includegraphics[width=\textwidth]{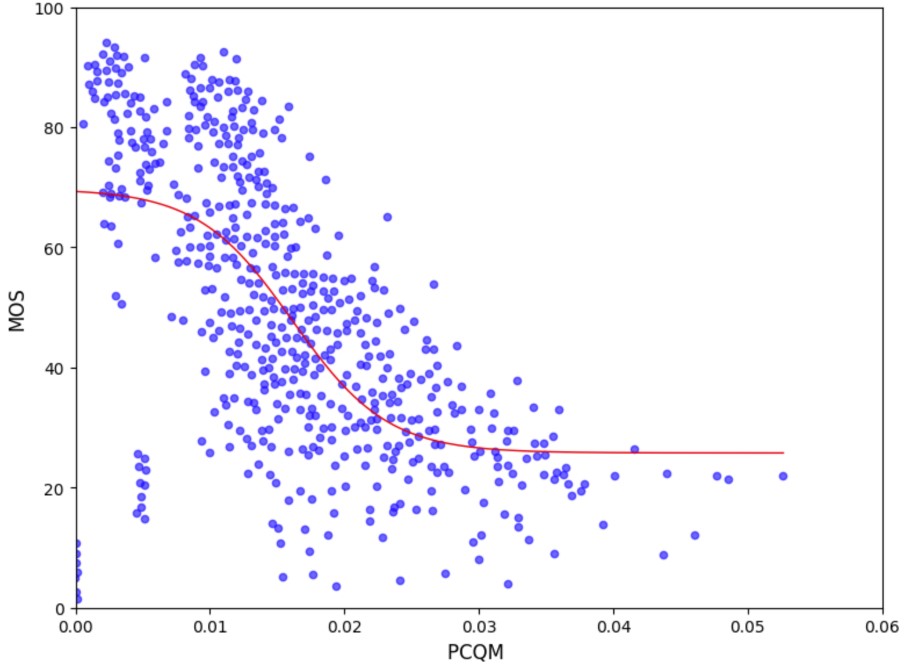}
     \end{subfigure}
    \hfill
     \begin{subfigure}{0.23\textwidth}
         \centering
         \captionsetup{justification=centering}
         \includegraphics[width=\textwidth]{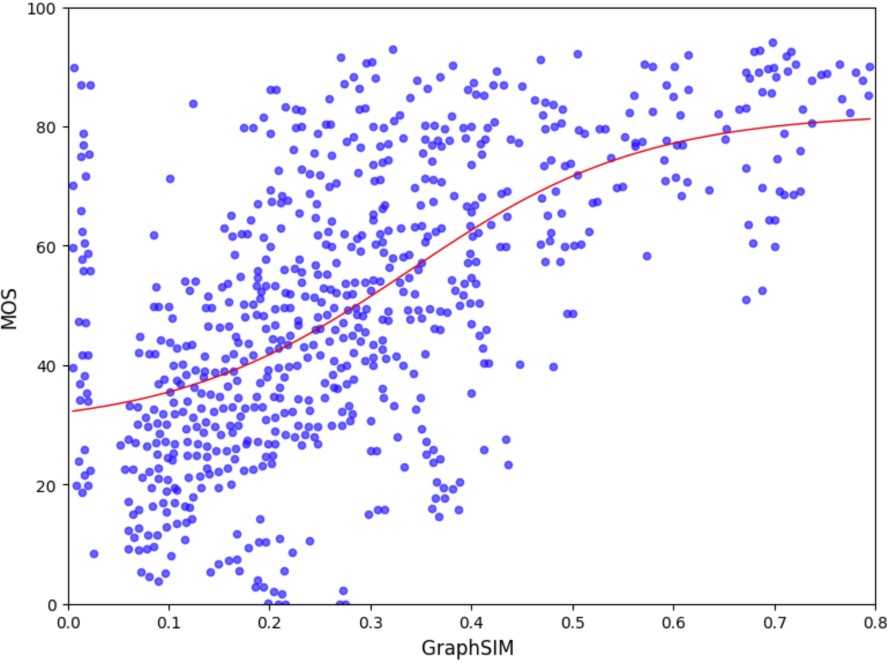}
     \end{subfigure}
     \hfill
     \begin{subfigure}{0.23\textwidth}
         \centering
         \captionsetup{justification=centering}
         \includegraphics[width=\textwidth]{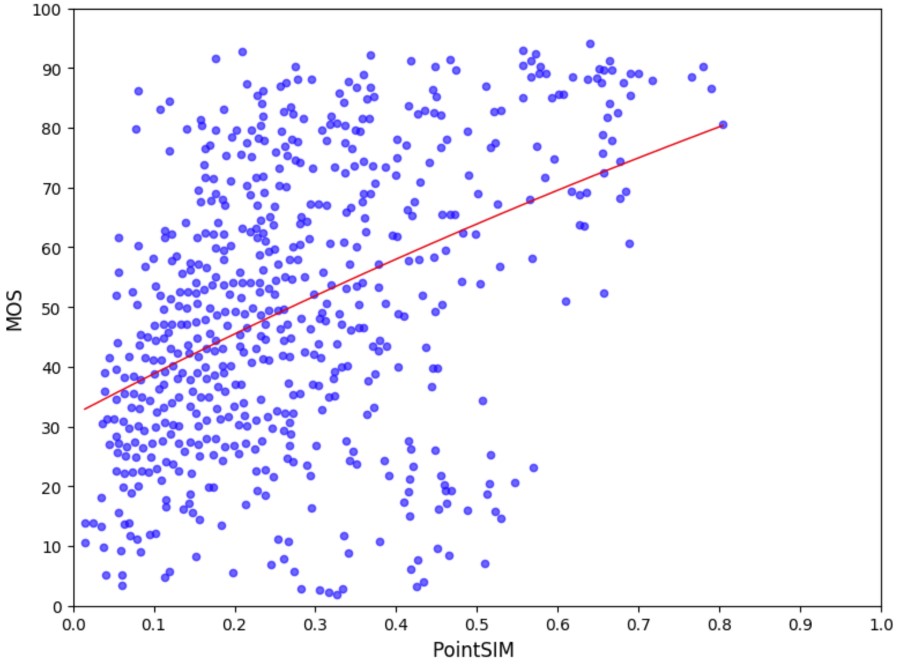}
     \end{subfigure}
    \hfill
     \begin{subfigure}{0.23\textwidth}
         \centering
         \captionsetup{justification=centering}
         \includegraphics[width=\textwidth]{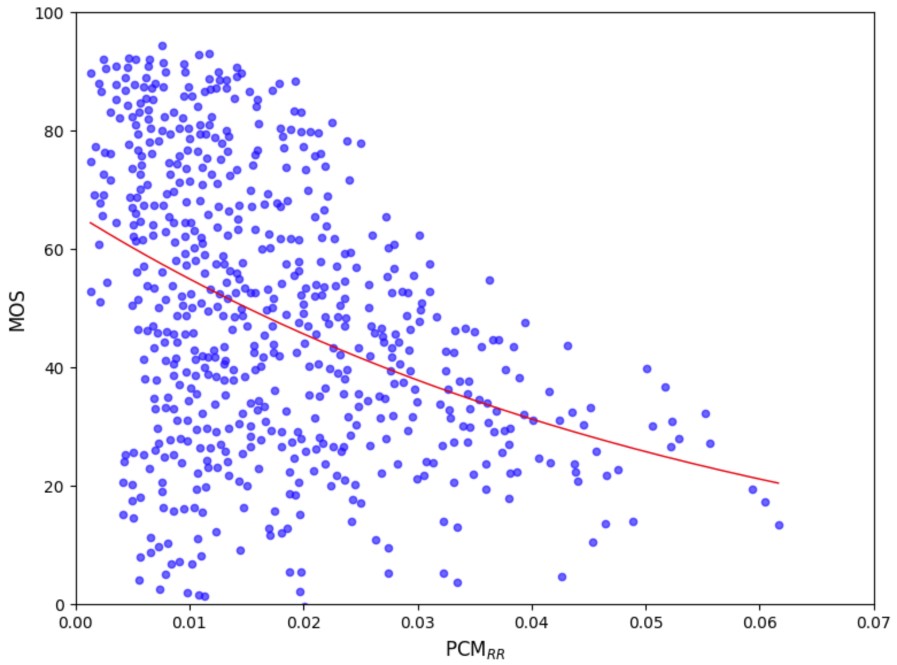}
     \end{subfigure}
     \hfill
     \begin{subfigure}{0.23\textwidth}
         \centering
         \captionsetup{justification=centering}
         \includegraphics[width=\textwidth]{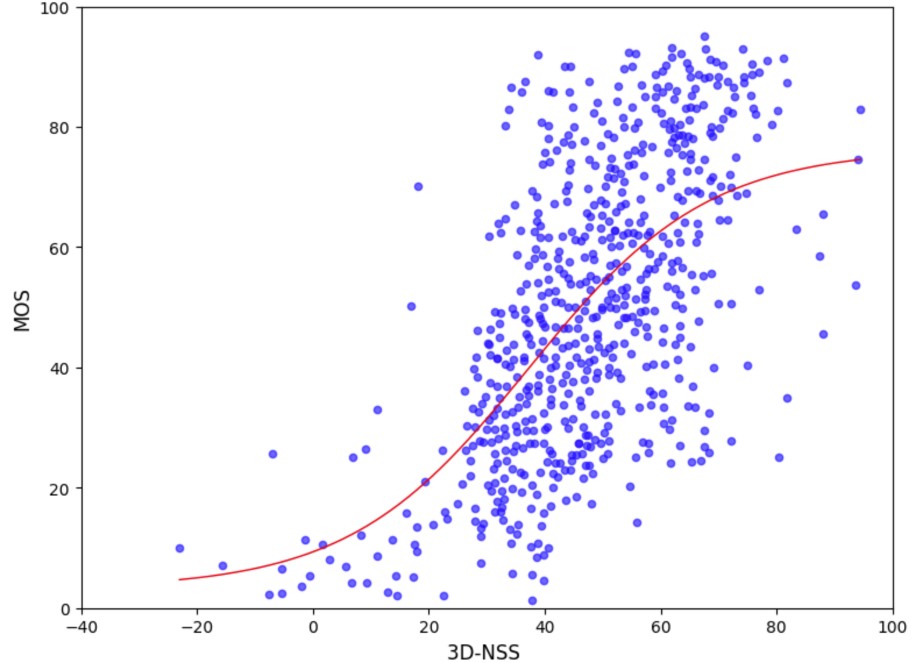}
     \end{subfigure}
     \hfill
     \begin{subfigure}{0.23\textwidth}
         \centering
         \captionsetup{justification=centering}
         \includegraphics[width=\textwidth]{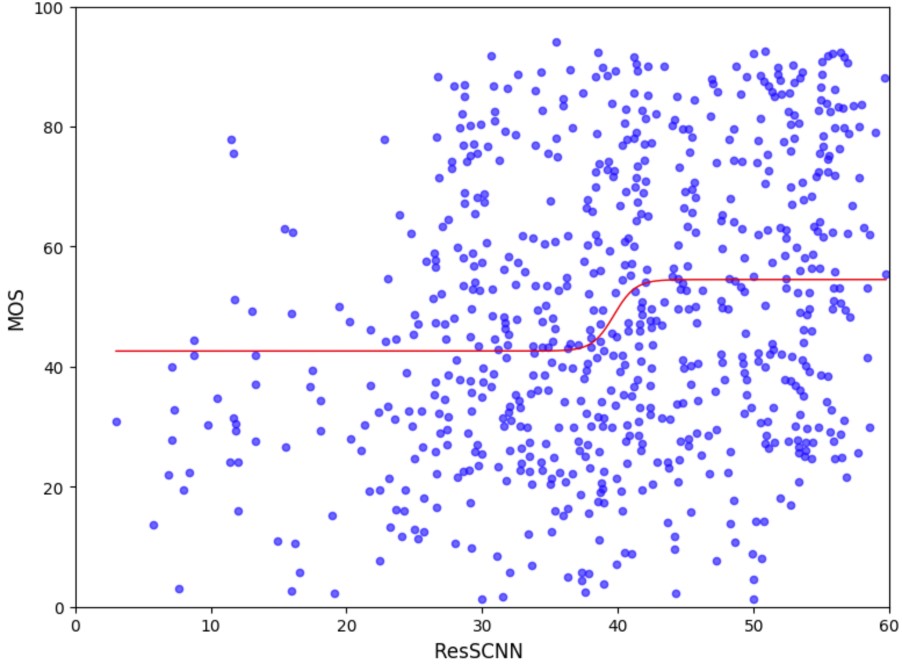}
     \end{subfigure}
     \hfill
     \begin{subfigure}{0.23\textwidth}
         \centering
         \captionsetup{justification=centering}
         \includegraphics[width=\textwidth]{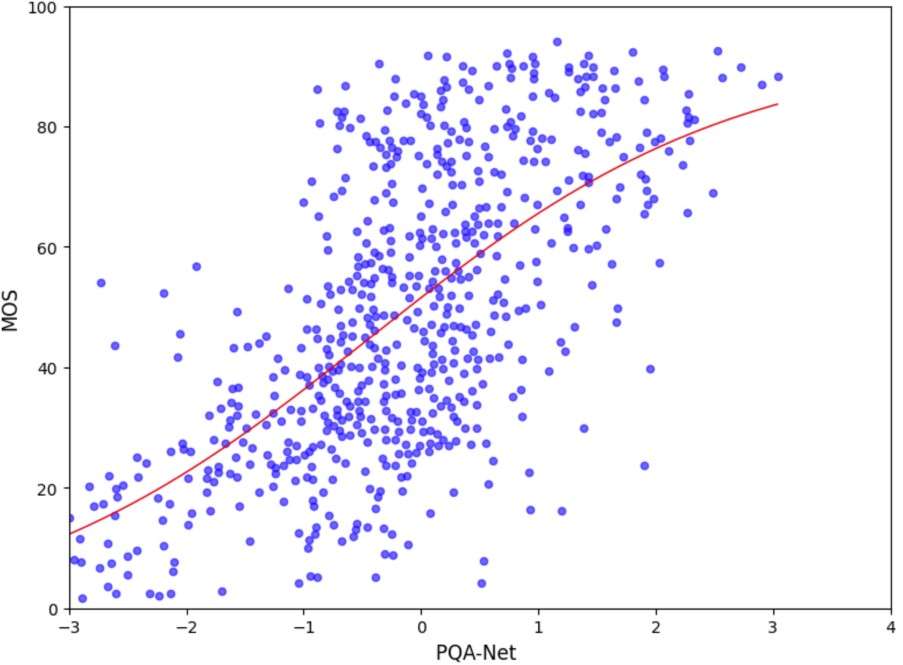}
     \end{subfigure}
     \hfill
     \begin{subfigure}{0.23\textwidth}
         \centering
         \captionsetup{justification=centering}
         \includegraphics[width=\textwidth]{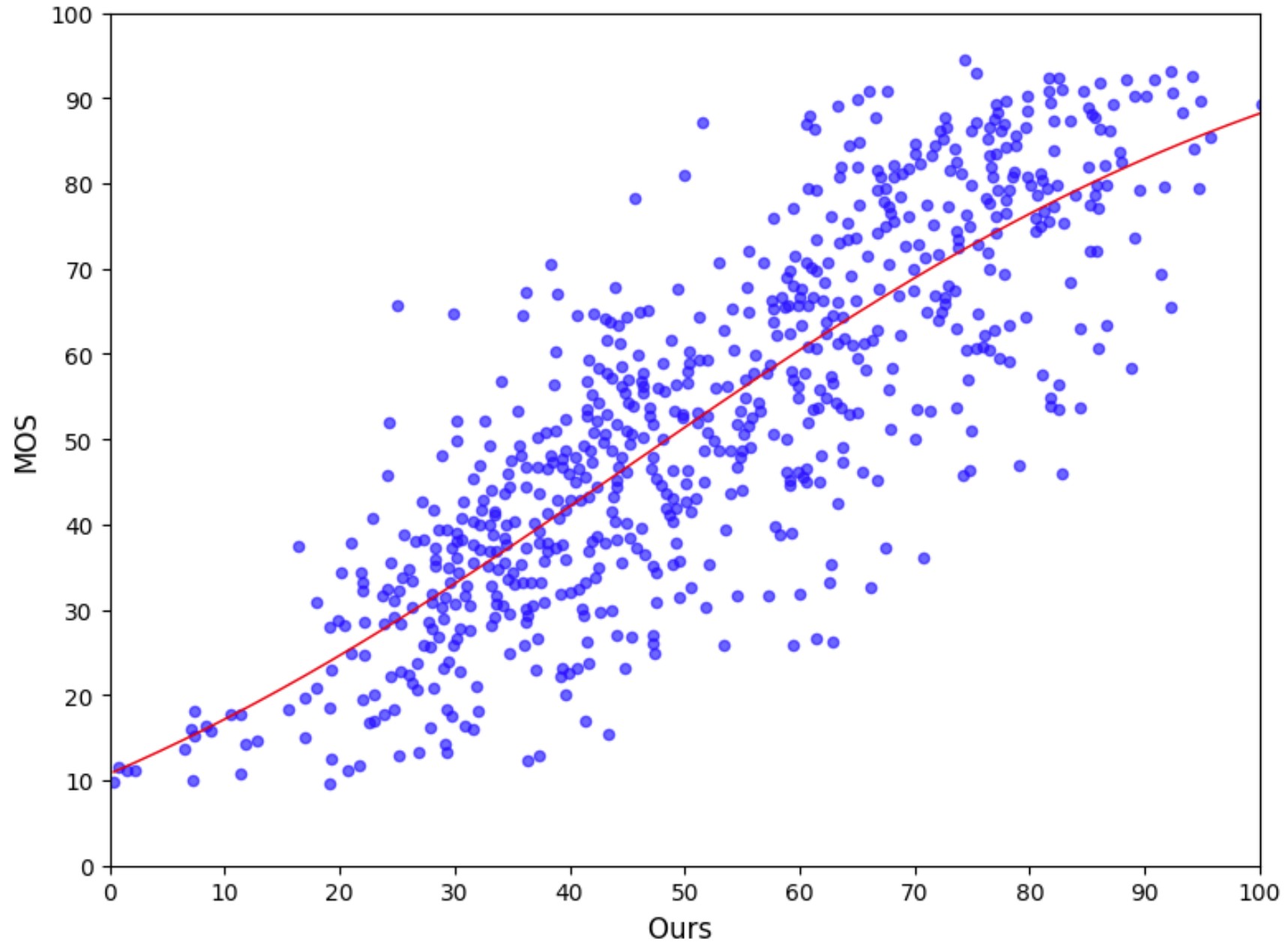}
     \end{subfigure}
     \hfill
     \begin{subfigure}{1\textwidth}
         \centering
         \captionsetup{justification=centering}
         \vspace{0.5cm}
         {(b) WPC}
     \end{subfigure}\\
     \caption{{Scatter plot between the objective scores and subjective MOS for the SJTU-PCQA (a) and WPC (b) databases. The red curve indicates the estimated no-linear logistic function.}.}
    \label{fig:plot-object-mos}
\end{figure*}

To visually illustrate the precision and consistency of our approach, Figure \ref{fig:plot-object-mos} presents scatter plots comparing the objective prediction scores with the subjective MOSs. The y-axis represents the subjective MOSs, while the x-axis shows the predicted scores. The red curves in the figure, obtained via non-linear fitting, help assess the correlation between the predicted and subjective scores. A curve closer to the diagonal line and a denser scatter distribution typically indicate better prediction performance. As seen in Figure \ref{fig:plot-object-mos}, our approach’s objective scores exhibit a stronger correlation with human judgments than competing methods, as evidenced by the closer alignment of the points with the fitted curve.
\begin{table}[!ht]
    \centering
    \caption{Performance comparison methods with our method.}
    \renewcommand{\arraystretch}{1.3} 
    \begin{tabular}{c|c|c|c}
        \hline
        \multirow{1}{*}{Ref} & \multirow{1}{*}{Type} & \multirow{1}{*}{Index} & \multirow{1}{*}{Metric} \\
        \hline
        \multirow{15}{*}{FR} & \multirow{11}{*}{PC-Based} &M1& PSNR$_{mse,p2po}$\cite{mekuria2016evaluation} \\
        & &M2& PSNR$_{mse,p2pl}$\cite{tian2017geometric}  \\
        & &M3& PSNR$_{hf,p2po}$ \cite{mekuria2016evaluation}\\
        & &M4& PSNR$_{hf,p2pl}$ \cite{tian2017geometric}\\
        & &M5& AS$_{mean}$ \cite{alexiou2018point}\\
        & &M6& AS$_{rms}$ \cite{alexiou2018point} \\
        & &M7& AS$_{mse}$ \cite{alexiou2018point} \\
        & &M8& PSNR$_{Y}$ \cite{mekuria2017performance}\\
        & &M9& PCQM \cite{meynet2020pcqm} \\
        & &M10& PointSSIM \cite{alexiou2020towards}\\
        & &M11& GraphSIM \cite{diniz2020local}\\
        \cline{2-4}
        & \multirow{4}{*}{Projection-Based} &M12& SSIM \cite{wang2004image}\\
        & &M13& MS-SSIM \cite{wang2003multiscale} \\
        & &M14& IW-SSIM \cite{wang2010information}\\
        & &M15& VIFP \cite{sheikh2006image} \\
        \hline
        \multirow{2}{*}{RR} & \multirow{1}{*}{PC-Based} &M16& PCMRR \cite{viola2020reduced}\\
        \cline{2-4}
        & \multirow{1}{*}{Projection-Based} &M17& RR-CAP \cite{zhou2023reduced} \\
        \hline
        \multirow{8}{*}{NR} & \multirow{3}{*}{PC-Based} &M18& 3D-NSS \cite{zhang2022no} \\
        & &M19& GPA-Net \cite{shan2023gpa}\\
        & &M20& ResSCNN \cite{liu2023point} \\
        \cline{2-4}
        & \multirow{4}{*}{Projection-Based} &M21& PQA-Net \cite{liu2021pqa}\\
        & &M22& IT-PCQA \cite{yang2022no} \\
        & &M23& GMS-3DQA \cite{zhang2024gms}\\
        & &M24& MM-PCQA \cite{zhang2022mm}\\
        \hline
    \end{tabular}
    
    \label{tab-metric-comparaison}
\end{table}

Given that point cloud quality databases contain diverse content and distortion types, it is essential to evaluate current PCQA methods across different point cloud content and distortion categories. Experimental results for the SJTU-PCQA and WPC databases are summarized in Tables \ref{tab-dist_sjtu} and \ref{tab-dist_wpc}. The meaning of the indices M1 through M25 is outlined in Table \ref{tab-metric-comparaison}.

For the SJTU-PCQA dataset (Table \ref{tab-dist_sjtu}), subsets such as ULB Unicorn and Longdress achieved the highest SRCC rankings. Notably, our method achieved top SRCC scores of 0.9166 and 0.9500, respectively. Subsets like Statue and Soldier also showed strong performance with SRCC scores of 0.9333 and 0.9123, respectively. On the other hand, subsets such as Romanoillamp and Shiva faced challenges, with lower SRCC values across all metrics. Nonetheless, our method achieved notable SRCC scores of 0.8667 for Romanoillamp and 0.8968 for Shiva, outperforming other methods like ResSCNN (M20), which obtained an SRCC score of 0.6193 for Romanoillamp.

Looking at different distortion types, our method performed well across various distortions, with SRCC values of 0.7727 for OT (Octree), 0.8181 for CN (color noise), and 0.8857 for DS (Downscaling). Additionally, our method achieved an SRCC of 0.9181 for the combined D+C (Downscaling and Color noise) distortion and 0.9200 for D+G (Downscaling and Geometry Gaussian noise). These results demonstrate the effectiveness of our method across a range of distortion types.
\begin{table*}[!ht]
\centering
\caption{The SRCC performance evaluation of current PCQA metrics is conducted on the SJTU-PCQA database based on different point cloud distortion types. Absolute SRCC values are used to enhance clarity in comparisons. The top three values in the SRCC criteria are highlighted in \textcolor{red}{red}, \textcolor{blue}{blue}, and \textcolor{green}{green} for first, second, and third place, respectively.}
\renewcommand{\arraystretch}{1.5} 
\resizebox{\textwidth}{!}{
\begin{tabular}{cc|c|c|c|c|c|c|c|c|c|c|c|c|c|c|c|c|c|c|c}
\hline   
\multicolumn{2}{c|}{\multirow{2}{*}{\textbf{Subset}}} & \multicolumn{13}{c|}{FR} & \multicolumn{1}{c|}{RR} & \multicolumn{5}{c}{NR}  \\ 
\cline{3-21}
                         && M1 \cite{mekuria2016evaluation}  & M2 \cite{tian2017geometric}  & M3 \cite{mekuria2016evaluation} & M4 \cite{tian2017geometric} & M6  \cite{alexiou2018point}& M8  \cite{mekuria2017performance}& M9  \cite{meynet2020pcqm}& M10 \cite{alexiou2020towards} & M11 \cite{diniz2020local} & M12 \cite{wang2004image} & M13  \cite{wang2003multiscale} & M14 \cite{wang2010information} & M15 \cite{sheikh2006image} & M16 \cite{viola2020reduced} & M18 \cite{zhang2022no} & M20 \cite{liu2023point} & M21 \cite{liu2021pqa} & M22 \cite{yang2022no} & {Ours} \\ \hline

   \multirow{9}{*}{\rotatebox{90}{\textbf{Content}}}
                         & \multicolumn{1}{|c|}{Hhi} & 0.6526 & 0.5150 & 0.7443 & 0.6785 & 0.5012 & 0.8242 & 0.7524 & 0.7010 & \textcolor{blue}{0.9028} & 0.8409 & 0.8658 & \textcolor{green}{0.8773} & 0.8462 & 0.4785 & 0.7394 & 0.8240 & 0.8409 & 0.7577 & \textcolor{red}{0.9036} \\ 
                         &\multicolumn{1}{|c|} {Longdress} & 0.6640 & 0.6437 & 0.7885 & 0.7096 & 0.5704 & \textcolor{green}{0.9326} & 0.8896 & 0.8608 & \textcolor{blue}{0.9499} & 0.9245 & 0.9191 & 0.8710 & 0.8976 & 0.6474 & 0.9005 & 0.8650 & 0.9245 & 0.8243 & \textcolor{red}{0.9500} \\
                         & \multicolumn{1}{|c|}{Loot} & 0.6738 & 0.6405 & 0.7447 & 0.6391 & 0.4817 & 0.7875 & 0.8426 & 0.7299 & \textcolor{blue}{0.8868} & 0.8693 & 0.8809 & \textcolor{green}{0.8846} & 0.8619 & 0.6770 & \textcolor{red}{0.8890} & 0.8780 & 0.8693 & 0.8778 & 0.8540 \\ 
                        & \multicolumn{1}{|c|}{Redandblack} & 0.6196 & 0.5943 & 0.7421 & 0.6819 & 0.5799 & 0.7478 & 0.8024 & 0.6670 & 0.8702 & 0.8603 & 0.8718 & \textcolor{blue}{0.8911} & \textcolor{green}{0.8885} & 0.6506 & 0.8647 & 0.8003 & 0.8603 & 0.8557 & \textcolor{red}{0.8999} \\ 
                         & \multicolumn{1}{|c|}{Romanoillamp} & 0.4247 & 0.3617 & 0.7457 & 0.6032 & 0.6022 & 0.4278 & 0.5145 & 0.5150 & \textcolor{blue}{0.8525} & 0.7509 & 0.7869 & \textcolor{green}{0.7939} & 0.7882 & 0.6044 & 0.6885 & 0.6193 & 0.7509 & 0.7248 & \textcolor{red}{0.8667} \\
                         & \multicolumn{1}{|c|}{Shiva} & 0.4129 & 0.4074 & 0.1168 & 0.2689 & 0.7057 & 0.8375 & 0.8060 & 0.7896 & 0.8595 & \textcolor{red}{0.8968} & \textcolor{blue}{0.8914} & 0.8744 & \textcolor{green}{0.8903} & 0.4884 & 0.8198 & 0.8599 & \textcolor{red}{0.8968} & 0.8243 & {0.8834} \\ 
                         & \multicolumn{1}{|c|}{Soldier} & 0.6781 & 0.6478 & 0.7493 & 0.6329 & 0.5404 & 0.8336 & 0.8684 & 0.7718 & \textcolor{blue}{0.9118} & 0.8917 & 0.8843 & 0.8843 & 0.8744 & 0.5809 & 0.8731 & \textcolor{red}{0.9123} & 0.8917 & 0.8050 & \textcolor{green}{0.9000} \\ 
                         & \multicolumn{1}{|c|}{Statue} & 0.5678 & 0.5362 & 0.5883 & 0.5652 & 0.6291 & 0.8241 & 0.7483 & 0.7391 & 0.8744 & 0.8578 & 0.8663 & 0.8428 & 0.8637 & 0.4181 & 0.8520 & \textcolor{blue}{0.9002} & 0.8578 & \textcolor{green}{0.8757} & \textcolor{red}{0.9333} \\
                         & \multicolumn{1}{|c|}{ULB Unicorn} & 0.7085 & 0.6082 & 0.8500 & 0.8081 & 0.4773 & 0.8687 & 0.7496 & 0.5715 & 0.8597 & \textcolor{green}{0.9084} & 0.8981 & 0.8548 & 0.8514 & 0.5148 & 0.4101 & 0.8364 & \textcolor{green}{0.9084} & \textcolor{blue}{0.9129} & \textcolor{red}{0.9166} \\ 
                         
                         \hline
\multirow{7}{*}{\rotatebox{90}{\textbf{Distortion}}}               
                        & \multicolumn{1}{|c|}{OT}  & 0.4407 & 0.4407 & 0.3788 & 0.3524 & 0.5210 & 0.3068 & 0.6495 & \textcolor{blue}{0.7108} & \textcolor{green}{0.7049} & 0.2198 & 0.2712 & 0.3382 & 0.3743 & 0.1800 & 0.4068 & 0.1683 & 0.0883 & 0.0189 & \textcolor{red}{0.7727} \\ 
                         & \multicolumn{1}{|c|}{CN} & NaN & NaN & NaN & NaN & NaN & 0.5588 & 0.6070 & \textcolor{green}{0.7660} & \textcolor{blue}{0.7779} & 0.6283 & 0.6453 & 0.7531 & 0.7429 & 0.7157 & 0.1480 & 0.2265 & 0.5507 & 0.0655 & \textcolor{red}{0.8181} \\ 
                         & \multicolumn{1}{|c|}{DS} & 0.4495 & 0.4489 & 0.6847 & 0.3286 & 0.3653 & 0.4697 & 0.6990 & \textcolor{green}{0.8500} & \textcolor{blue}{0.8654} & 0.3246 & 0.4718 & 0.4535 & 0.4546 & 0.1489 & 0.5051 & 0.4292 & 0.2958 & 0.0556 & \textcolor{red}{0.8857} \\ 
                         & \multicolumn{1}{|c|}{D+C} & 0.5735 & 0.5979 & 0.7619 & 0.7499 & 0.4025 & 0.7397 & \textcolor{green}{0.8014} & 0.7449 & \textcolor{blue}{0.8846} & 0.5062 & 0.6281 & 0.6661 & 0.6932 & 0.6120 & 0.5895 & 0.5158 & 0.4899 & 0.0468 & \textcolor{red}{0.9181} \\
                         & \multicolumn{1}{|c|}{D+G} & 0.6779 & 0.7058 & 0.7423 & 0.7196 & \textcolor{green}{0.8915} & 0.5413 & 0.7476 & \textcolor{blue}{0.9288} & 0.8833 & 0.6920 & 0.7589 & 0.8222 & 0.7989 & 0.7439 & 0.7442 & 0.5263 & 0.5033 & 0.0411 & \textcolor{red}{0.9545} \\
                         & \multicolumn{1}{|c|}{GGN} &0.7008	&0.7144	&0.7453	&0.7328	&\textcolor{blue}{0.9376}	&0.5727	&0.7143	&0.9027	&\textcolor{green}{0.9064}	&0.7436	&0.7783	&0.8324	&0.8436	&0.7813	&0.8435	&0.4497	&0.3771	&0.0798	&\textcolor{red}{0.9636} \\
                         & \multicolumn{1}{|c|}{C+G} &0.7577	&0.7758	&0.8205	&0.8025	&0.9241	&0.6692	&0.7078	&0.7991	&\textcolor{blue}{0.9334}	&0.7307	&0.7948	&0.8406	&0.8463	&0.8329	&\textcolor{green}{0.8645}	&0.5523	&0.6137	&0.1044	&\textcolor{red}{0.9454} \\ 
                         \hline
\end{tabular}}
\label{tab-dist_sjtu}
\end{table*}

On the WPC dataset (Table \ref{tab-dist_wpc}), point clouds such as Biscuits, Honeydew Melon, and Pen Container achieved the highest SRCC scores on most metrics, indicating they are reliable benchmarks for quality assessment. Biscuits, in particular, performed exceptionally well with an SRCC score above 0.95, reflecting strong quality alignment. Honeydew Melon also showed robust resistance to distortions, with scores similar to other high-performing methods. In contrast, point clouds such as Cake, Cauliflower, and Pineapple had lower SRCC scores, making quality assessment more challenging. These cases indicate potential weaknesses in the metrics, suggesting a need for further refinement to capture quality variation more effectively.
\begin{table*}[!ht]
\centering
\caption{The SRCC performance evaluation of current PCQA metrics is conducted on the WPC database based on different point cloud distortion types. Absolute SRCC values are used to enhance clarity in comparisons. The top three values in the SRCC criteria are highlighted in \textcolor{red}{red}, \textcolor{blue}{blue}, and \textcolor{green}{green} for first, second, and third place, respectively.}
\renewcommand{\arraystretch}{1.5} 
\resizebox{\textwidth}{!}{
\begin{tabular}{cc|c|c|c|c|c|c|c|c|c|c|c|c|c|c|c|c|c|c|c}
\hline   
\multicolumn{2}{c|}{\multirow{2}{*}{\textbf{Subset}}} & \multicolumn{13}{c|}{FR} & \multicolumn{1}{c|}{RR} & \multicolumn{5}{c}{NR}  \\ 
\cline{3-21}
                         && M1 \cite{mekuria2016evaluation}  & M2 \cite{tian2017geometric}  & M3 \cite{mekuria2016evaluation} & M4 \cite{tian2017geometric} & M6  \cite{alexiou2018point}& M8  \cite{mekuria2017performance}& M9  \cite{meynet2020pcqm}& M10 \cite{alexiou2020towards} & M11 \cite{diniz2020local} & M12 \cite{wang2004image} & M13  \cite{wang2003multiscale} & M14 \cite{wang2010information} & M15 \cite{sheikh2006image} & M16 \cite{viola2020reduced} & M18 \cite{zhang2022no} & M20 \cite{liu2023point} & M21 \cite{liu2021pqa} & M22 \cite{yang2022no} & {Ours} \\ \hline

   \multirow{20}{*}{\rotatebox{90}{\textbf{Content}}}   
                        & \multicolumn{1}{|c|}{Bag}	&0.6669	&0.5751	&0.4363	&0.4365	&0.4325	&\textcolor{blue}{0.8051}	&0.5955	&0.4829	&0.7164	&0.7300	&0.7584	&0.7309	&0.7093	&0.6069	&\textcolor{green}{0.7731}	&0.1603	&0.3504	&0.6174	&\textcolor{red}{0.8263} \\ 
                         & \multicolumn{1}{|c|}{Banana}	&0.6471	&0.5691	&0.1933	&0.2033	&0.3147	&0.6211	&0.4649	&0.2202	&0.5045	&\textcolor{blue}{0.8011}	&0.7677	&\textcolor{green}{0.7790}	&0.7771	&0.5287	&0.6524	&0.2475	&0.6949	&0.2485	&\textcolor{red}{0.8333} \\
                         & \multicolumn{1}{|c|}{Biscuits}	&0.5252	&0.4160	&0.3085	&0.3368	&0.3505	&0.7764	&0.6245	&0.5816	&0.7198	&\textcolor{green}{0.9173}	&\textcolor{blue}{0.9500}	&0.7992	&0.7416	&0.4310	&0.6645	&0.4765	&0.6147	&0.3570	&\textcolor{red}{0.9523} \\ 
                         & \multicolumn{1}{|c|}{Cake}	&0.3074	&0.1798	&0.1724	&0.1796	&0.0609	&0.5180	&0.4566	&0.3177	&0.4251	&\textcolor{green}{0.7390}	&\textcolor{blue}{0.7691}&0.6534	&0.6477	&0.3070	&0.4547	&0.4467	&0.5835	&0.7300	&\textcolor{red}{0.8095} \\ 
                         & \multicolumn{1}{|c|}{Cauliflower}	&0.3501	&0.2058	&0.0918	&0.1653	&0.1781	&0.5927	&0.4903	&0.4237	&0.5529	&0.8004	&\textcolor{red}{0.8608}	&\textcolor{green}{0.8182}	&0.7008	&0.4187	&0.5517	&0.5095	&0.6238	&0.0593	&\textcolor{blue}{0.8572} \\
                         &\multicolumn{1}{|c|} {Flowerpot}	&0.6509	&0.5298	&0.4348	&0.4515	&0.3629	&0.6385	&0.5875	&0.3784	&0.6609	&0.8303	&\textcolor{blue}{0.9066}	&\textcolor{green}{0.9047}	&0.8954	&0.0477	&0.6958	&0.4900	&0.2357	&0.8127	&\textcolor{red}{0.9286} \\
                         & \multicolumn{1}{|c|}{GlassesCase}	&0.5845	&0.4390	&0.2020	&0.3238	&0.4288	&\textcolor{blue}{0.7826}	&0.5861	&0.5258	&0.6546	&0.7617	&0.7577	&0.7304	&0.7459	&0.3883	&0.4790	&0.2003	&0.7674	&\textcolor{green}{0.7750}	&\textcolor{red}{0.7857 }\\
                         & \multicolumn{1}{|c|}{HoneydewMelon}	&0.4890	&0.3299	&0.2768	&0.2300	&0.3228	&0.6740	&0.4500	&0.5609	&0.7248	&0.8549	&\textcolor{blue}{0.8917}	&\textcolor{red}{0.9180}	&0.8279	&0.5742	&0.7229	&0.4026	&0.7418	&0.7352	&\textcolor{green}{0.8810} \\ 
                         & \multicolumn{1}{|c|}{House}	&0.5866	&0.4483	&0.3429	&0.3434	&0.4522	&0.7798	&0.5880	&0.5590	&0.7373	&\textcolor{green}{0.7788}	&0.7793	&0.7357	&0.7200	&0.4905	&0.7646	&0.4780	&\textcolor{blue}{0.8668}	&0.4201	&\textcolor{red}{0.9048} \\ 
                         & \multicolumn{1}{|c|}{Litchi}	&0.5109	&0.4291	&0.3478	&0.3204	&0.3554	&0.7027	&0.5965	&0.6422	&0.6958	&0.7748	&\textcolor{red}{0.8623}	&0.7496	&0.7018	&0.4839	&\textcolor{green}{0.8113}	&0.1994	&0.7207	&0.0868	&\textcolor{blue}{0.8456} \\
                         & \multicolumn{1}{|c|}{Mushroom}	&0.6396	&0.5156	&0.3486	&0.3105	&0.2911	&0.6550	&0.5725	&0.5443	&0.6802	&0.7821	&\textcolor{red}{0.8781}	&\textcolor{green}{0.8160}	&0.7897	&0.2556	&0.8153	&0.0754	&0.5835	&0.3570	&\textcolor{blue}{0.8742} \\
                         & \multicolumn{1}{|c|}{PenContainer} 	&0.7720	&0.6688	&0.2159	&0.3635	&0.5465	&0.7328	&0.6394	&0.5948	&0.8250	&\textcolor{red}{0.8954}	&\textcolor{green}{0.8758}	&0.8485	&0.8397	&0.6830	&0.7809	&0.5676	&0.6470	&0.7859	&\textcolor{blue}{0.8854} \\
                         & \multicolumn{1}{|c|}{Pineapple}	&0.3777	&0.2785	&0.1376	&0.1831	&0.2155	&0.7217	&0.6427	&0.5386	&0.6401	&\textcolor{green}{0.7307}	&\textcolor{red}{0.7805}	&0.5856	&0.6441	&0.4011	&0.6074	&0.5275	&0.6318	&0.5913	&\textcolor{blue}{0.7619} \\
                         & \multicolumn{1}{|c|}{PingpongBat}	&0.5924	&0.4984	&0.4958	&0.4357	&0.4521	&0.5428	&0.5783	&0.6051	&0.7697	&\textcolor{green}{0.8054}	&\textcolor{red}{0.8812}	&0.7570	&0.7539	&0.5092	&0.6935	&0.3518	&0.6358	&0.4737	&\textcolor{blue}{0.8571} \\
                         & \multicolumn{1}{|c|}{PuerTea}	&0.6069	&0.4746	&0.1173	&0.0384	&0.4734	&0.7639	&0.5685	&0.4139	&0.7999	&\textcolor{blue}{0.8917}	&\textcolor{green}{0.8668}	&0.8359	&0.7866	&0.4308	&0.4763	&0.1456	&0.7359	&0.5467	&\textcolor{red}{0.8982} \\
                         & \multicolumn{1}{|c|}{Pumpkin}	&0.4947	&0.3423	&0.3092	&0.3068	&0.3220	&0.6901	&0.5934	&0.5699	&0.6517	&\textcolor{red}{0.9111}	&0.8156	&\textcolor{blue}{0.9042}	&\textcolor{green}{0.8976} &0.3241	&0.5768	&0.4052	&0.7857	&0.5536	&0.8917 \\
                         & \multicolumn{1}{|c|}{Ship}	&0.7464	&0.6267	&0.3404	&0.5158	&0.4943	&0.7786	&0.5434	&0.4488	&0.7558	&\textcolor{blue}{0.8973}	&\textcolor{green}{0.8578}	&0.8340	&0.8013	&0.4400	&0.6935	&0.6612	&0.5349	&0.3777	&\textcolor{red}{0.9025} \\
                         & \multicolumn{1}{|c|}{Statue}	&0.8040	&0.6707	&0.2450	&0.4487	&0.4900	&0.7001	&0.5714	&0.5085	&0.7390	&0.8985	&\textcolor{red}{0.9372}	&\textcolor{green}{0.9099}	&0.8950	&0.1811	&0.6368	&0.5782	&0.3762	&0.4976	&\textcolor{blue}{0.9285} \\
                         & \multicolumn{1}{|c|}{Stone}	&0.6219	&0.5129	&0.3551	&0.3424	&0.3649	&0.7115	&0.6475	&0.6126	&0.1920	&0.8426	&\textcolor{red}{0.8881}	&\textcolor{green}{0.8587}	&0.8196	&0.3632	&0.6968	&0.2122	&0.8234	&0.1790	&\textcolor{red}{0.8639} \\
                         & \multicolumn{1}{|c|}{ToolBox}	&0.3937	&0.2969	&0.1972	&0.1884	&0.2984	&\textcolor{blue}{0.8706}	&0.6304	&0.4927	&0.7935	&0.7821	&0.8255	&0.8056	&0.7411	&0.5239	&0.5806	&0.5026	&\textcolor{green}{0.8653}	&0.4694	&\textcolor{red}{0.8858}\\
                         \hline
\multirow{5}{*}{\rotatebox{90}{\textbf{Distortion}}}               
                        & \multicolumn{1}{|c|}{Downsampling}	&0.4815	&0.3251	&0.5356	&0.4879	&0.2465	&0.5542	&0.4537	&0.8319	&0.7903	&\textcolor{blue}{0.8234}	&0.8834	&0.8822	&\textcolor{green}{0.8828}	&0.7407	&0.7508	&0.2899	&0.7234	&0.3327	&\textcolor{red}{0.8951} \\ 
                         & \multicolumn{1}{|c|}{Gaussian noise}	&0.6155	&0.6194	&0.6149	&0.6150	&0.6844	&0.7644	&\textcolor{green}{0.8775}	&0.5844	&0.7469	&0.6264	&0.7118	&0.8560	&\textcolor{blue}{0.8847}	&0.7762	&0.7460	&0.5459	&0.7938	&0.1718	&\textcolor{red}{0.9133} \\ 
                         & \multicolumn{1}{|c|}{G-PCC (T)}	&0.3451	&0.3568	&0.2811	&0.3085	&0.1342	&0.5916	&\textcolor{red}{0.7775}	&0.6745	&\textcolor{green}{0.7457}	&0.4669	&0.6042	&0.6742	&0.6304	&0.2702	&0.5947	&0.2531	&0.4710	&0.1987	&\textcolor{blue}{0.7652}\\ 
                         & \multicolumn{1}{|c|}{V-PCC}	&0.1602	&0.1992	&0.2051	&0.2370	&0.3877	&0.3203	&0.5534	&0.3546	&0.5989	&0.5141	&0.5812	&\textcolor{green}{0.7063}	&\textcolor{blue}{0.7410}	&0.2966	&0.3927	&0.1028	&0.0045	&0.0090	&\textcolor{red}{0.7750}\\
                         & \multicolumn{1}{|c|}{G-PCC (O)}	&NaN	&NaN	&NaN	&NaN	&0.0350	&0.8072	&\textcolor{blue}{0.8944}	&0.7917	&\textcolor{green}{0.8258}	&0.5290	&0.7214	&0.7128	&0.7116	&0.6468	&0.2891	&0.0247	&0.4204	&0.1180	&\textcolor{red}{0.9047} \\
                         
                         \hline
\end{tabular}}
\label{tab-dist_wpc}

\end{table*}

Overall, these results highlight that our approach consistently outperforms the other methods across the two databases, achieving the most top-three results and the highest number of best results overall. On the SJTU-PCQA dataset, our method ranked in the top three SRCC scores in 14 out of 16 subset experiments, securing the best performance in 13 of 16 experiments. On the WPC dataset, our method ranked among the top three SRCC performances in 24 out of 25 subset experiments, including 15 best performances. These results demonstrate the effectiveness of our method in various quality assessment scenarios.

In conclusion, our no-reference (NR) method performs exceptionally well, achieving first place in the number of best performances across all three databases. This demonstrates that our approach is highly effective in predicting the perceptual quality of point clouds and is robust to complex visual distortions.

\subsection{Ablation Study}
To evaluate the contributions and effectiveness of each perceptual feature, we conducted independent performance tests. By assessing each feature's performance in various combinations, we analyzed its individual contributions and the importance of the Graph Attention Fusion (GAF) mechanism in improving the accuracy of point cloud quality assessment. Without GAF, the final predicted quality score is obtained by averaging the scores. The results of the ablation study are presented in Table \ref{tab-ablation}, where "Curv," "Sal," and "LAB" represent curvature, saliency, and the LAB color space, respectively.

\begin{table*}[!ht]
    \centering
    \caption{Performance of the ablation study on the ICIP2020, WPC, and SJTU-PCQA databases. The top three values for PLCC and SRCC are highlighted in \textcolor{red}{red}, \textcolor{blue}{blue}, and \textcolor{green}{green} for first, second, and third place, respectively.}
    \renewcommand{\arraystretch}{1.5} 
    \resizebox{\textwidth}{!}{
    \begin{tabular}{c|cc|cc|cc}
        \hline
        \multirow{2}{*}{Feature} & \multicolumn{2}{c|}{SJTU-PCQA} & \multicolumn{2}{c|}{WPC} & \multicolumn{2}{c}{ICIP2020} \\
        \cline{2-7}
         & PLCC↑ & SRCC↑  & PLCC↑ & SRCC↑  & PLCC↑ & SRCC↑  \\
        \hline
        LAB & 0.7586 & 0.7660 & 0.7056 & 0.6859 & 0.8546 & 0.8787 \\
        CURV  & 0.7803 & 0.7726 & 0.6786 & 0.6456 & 0.9006 & 0.9151 \\
        SAL & 0.8550 & 0.8491 & 0.7395 & 0.7121 & 0.9263 & 0.9215 \\
        \hline
        LAB+CURV (without GAF)  & 0.7695 & 0.7693 & 0.6921 & 0.6658 & 0.8776 & 0.8969 \\
        LAB+CURV (with GAF)  & 0.8360 & 0.8487 & 0.7041 & 0.7034 & 0.9320 & 0.9272 \\
        LAB+SAL (without GAF)  & 0.8068 & 0.8076 & 0.7226 & 0.6990 & 0.8905 & 0.9001 \\
        LAB+SAL (with GAF)  & \textcolor{green}{0.8578} & \textcolor{green}{0.8494} & \textcolor{green}{0.7410} & \textcolor{green}{0.7206} & \textcolor{green}{0.9627} & \textcolor{blue}{0.9726} \\
        CURV+SAL (without GAF)  & 0.8177 & 0.8109 & 0.7091 & 0.6789 & 0.9135 & 0.9183 \\
        CURV+SAL (with GAF)  & \textcolor{blue}{0.8885} & \textcolor{blue}{0.8714} & \textcolor{blue}{0.7438} & \textcolor{blue}{0.7306} & \textcolor{blue}{0.9636} & \textcolor{green}{0.9665} \\
        LAB+CURV+SAL (without GAF)  & 0.7980 & 0.7959 & 0.7079 & 0.6812 & 0.8938 & 0.9051 \\
        LAB+CURV+SAL (with GAF)     & \textcolor{red}{0.9338} & \textcolor{red}{0.9118} & \textcolor{red}{0.8603} & \textcolor{red}{0.8688} & \textcolor{red}{0.9926} & \textcolor{red}{0.9999} \\
        \hline
    \end{tabular}}
    \label{tab-ablation}
\end{table*}

The LAB feature, which captures color information, demonstrates moderate performance across the three databases. Its highest performance is observed on the ICIP2020 database (PLCC = 0.8546, SRCC = 0.8787), followed by SJTU-PCQA (PLCC = 0.7586, SRCC = 0.7660) and WPC (PLCC = 0.7056, SRCC = 0.6859). While LAB is consistent, it appears insufficient to adequately represent perceptual quality on its own. In contrast, the CURV feature, which captures structural distortions, performs better, particularly on ICIP2020 (PLCC = 0.9006, SRCC = 0.9151). It also performs well on SJTU-PCQA (PLCC = 0.7803, SRCC = 0.7726) and WPC (PLCC = 0.6786, SRCC = 0.6456). The SAL (Saliency) feature is the most effective single descriptor, achieving the best scores across all datasets, especially on ICIP2020 (PLCC = 0.9263, SRCC = 0.9215). This highlights the importance of saliency in aligning with human perceptual judgment for point cloud quality assessment.

The results of combining features vary significantly depending on whether GAF is used. For instance, the combination of LAB+CURV without GAF yields moderate improvements, with the best results on ICIP2020 (PLCC = 0.8776, SRCC = 0.8969). However, incorporating GAF with this combination significantly enhances performance, as seen on ICIP2020 (PLCC = 0.9320, SRCC = 0.9272) and SJTU-PCQA (PLCC = 0.8360, SRCC = 0.8487). Similarly, LAB+SAL performs well without GAF, particularly on ICIP2020 (PLCC = 0.8905, SRCC = 0.9001), but GAF further boosts its effectiveness across all databases. For example, PLCC and SRCC improve to 0.8578 and 0.8494, respectively, on SJTU-PCQA. GAF also enhances the CURV+SAL combination, showing remarkable gains on SJTU-PCQA (PLCC = 0.8885, SRCC = 0.8714) and outstanding performance on ICIP2020 (PLCC = 0.9636, SRCC = 0.9665).

The most promising results are achieved with the full combination of LAB+CURV+SAL, especially when integrated with GAF. Without GAF, this combination performs well but not optimally, with ICIP2020 achieving PLCC = 0.8938 and SRCC = 0.9051, and SJTU-PCQA achieving PLCC = 0.7980 and SRCC = 0.7959. However, with GAF, this combination delivers exceptional results, achieving PLCC = 0.9338 and SRCC = 0.9118 on SJTU-PCQA, and PLCC = 0.8603 and SRCC = 0.8688 on WPC. On ICIP2020, it achieves near-perfect scores (PLCC = 0.9926, SRCC = 0.9999). These improvements underscore GAF's ability to enhance feature interactions and capture complex relationships among perceptual attributes, leading to superior generalization across diverse datasets.

The findings highlight the importance of combining perceptual features for robust performance and demonstrate the critical role of GAF in capturing complex distortions and achieving consistent quality predictions across datasets. Overall, the results emphasize the significance of integrating multiple features and leveraging GAF for accurate and reliable point cloud quality assessment.
\subsection{Cross-Database Evaluation}
To assess the generalization capability of the proposed approach, we conducted a cross-database evaluation. The experimental results are presented in Table \ref{tab_cross_data}.

\begin{table}[!ht]
   \begin{center} 
    \caption{Cross-database evaluation, where "WPC$\rightarrow$SJTU" indicates that the model is trained on the WPC dataset and validated on the SJTU-PCQA dataset. The best performance is highlighted in \textcolor{red}{red}.}
    \begin{tabular}{c|c c}
    \hline
    \multirow{2}{*}{Method} & \multicolumn{2}{c}{WPC$\rightarrow$SJTU} \\
    & PLCC & SRCC \\
    \hline
    3D-NSS \cite{zhang2022no} & 0.2344 & 0.1817 \\
    ResSCNN \cite{zhang2022no} & 0.4089 & 0.4031 \\
    PQA-net \cite{liu2021pqa} & 0.6102 & 0.5411 \\
    MM-PCQA \cite{zhang2022mm} & 0.7779 & 0.7693 \\
    Proposed method & \textcolor{red}{0.7928} & \textcolor{red}{0.8008} \\
    \hline
    \end{tabular}
    \label{tab_cross_data}
    \end{center}
\end{table}

Given the large size of the WPC database (740 samples), we focused on using this dataset for model training, while the SJTU-PCQA database (378 samples) was used for validation. Among the comparison models, 3D-NSS \cite{zhang2022no} yielded the lowest performance, with PLCC and SRCC values of 0.2344 and 0.1817, respectively. PQA-net \cite{liu2021pqa} showed improved performance, achieving PLCC = 0.6102 and SRCC = 0.5411. MM-PCQA \cite{zhang2022mm} further improved these metrics with PLCC = 0.7779 and SRCC = 0.7693. However, the proposed method outperformed all other models, achieving the highest values of PLCC = 0.7928 and SRCC = 0.8008. These results indicate that the proposed model demonstrates a stronger correlation with ground truth data, outperforming current NR-PCQA methods, including MM-PCQA, ResSCNN, PQA-net, and 3D-NSS. This highlights the superior generalization capability of our approach compared to other quality assessment methods.

\subsection{Computational Efficiency}

\begin{figure}[!ht]
\centering
\includegraphics[width=0.5\textwidth]{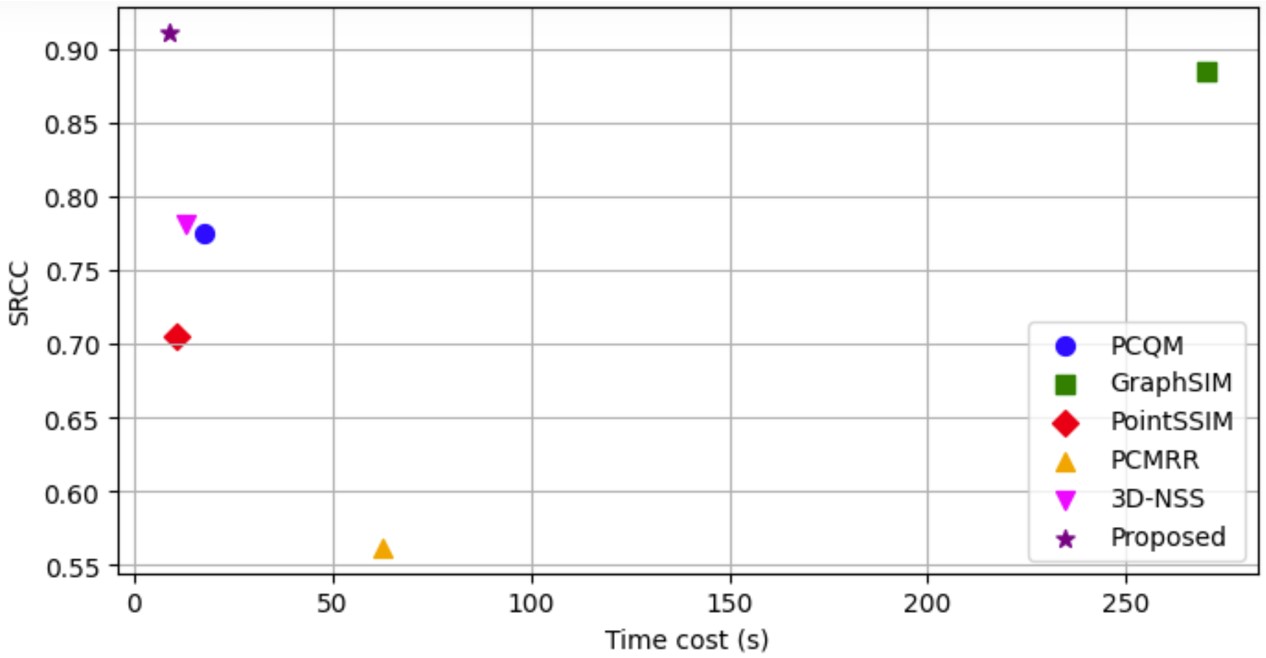}
\caption{Average computation time compared to SRCC results on the SJTU-PCQA database.}
\label{fig_time_cost}
\end{figure}

Given that the proposed approach operates directly on the 3D model, computational efficiency is a key consideration. To provide a meaningful comparison, we benchmarked our method against model-based techniques such as PCQM, PCMRR, GraphSIM, PointSSIM, and 3D-NSS. The corresponding time cost results, shown in Figure \ref{fig_time_cost}, clearly illustrate that the proposed method has a lower average time cost than all the benchmarked model-based methods, highlighting its relatively high computational efficiency. The experiments were conducted on a Windows-based computer equipped with an NVIDIA GeForce RTX 3060 laptop GPU, 32 GB of RAM, and an Intel (R) Core (TM) i7-11800H @ 2.30 GHz processor.

\section{Conclusion}

In This paper, we proposed Perceptual Clustering Weighted Graph (PCW-Graph) based method for No-Reference Point Cloud Quality Assessment (NR-PCQA). By combining a graph-based framework with a Graph Attention Fusion (GAF) network, our approach effectively captures distortions in point clouds, leading to accurate quality predictions. Results from the SJTU-PCQA, WPC, and ICIP2020 datasets show that our method consistently outperforms current techniques in aligning with human judgments, making it a reliable tool for real-world applications where reference models are unavailable.

Our approach, which incorporates perceptual features such as color, curvature, and saliency, ensures that the model efficiently detects distortions while maintaining computational efficiency. The weighted graph and GAF module work together to reveal the complex relationships between different features, enhancing the overall quality assessment. The model's ability to generalize across various datasets and types of distortions sets it apart from other NR-PCQA methods. The ablation study further confirms the importance of combining features and using GAF to improve prediction accuracy.

However, there are some limitations to consider. The method’s reliance on clustering techniques like K-means can be sensitive to the chosen number of clusters, which may affect performance in cases of irregular point distributions or noisy datasets. While efficient, processing large datasets with many features can increase computational demands, requiring optimization for large-scale applications. Additionally, the model’s effectiveness may decrease in scenarios where the importance of perceptual features varies significantly across different point cloud types.

Future work could focus on integrating multi-modal data through projections (e.g., 2D image projections), point-based features (e.g., LiDAR point clouds), and feature-based data (e.g., texture or color information). A graph foundation model, leveraging a large language model (LLM), could be employed to capture the complex relationships between these diverse data types. This approach would enhance the robustness and scalability of point cloud quality assessment, enabling better generalization across various distortion types and datasets.

\end{document}